\documentclass[runningheads]{llncs}
\usepackage{amssymb}
\usepackage{xcolor}
\usepackage{graphicx}
\usepackage{url}
\newcommand*{\captionsource}[2]{%
  \caption[{#1}]{%
    #1%
    \;
    \textbf{Source:} #2%
  }%
}

\begin{document}
%
\title{Black Box Model Explanations and the Expectation of Human Interpretation -  An Analyzes in the Context of Homicide Prediction}
%

 \author{José de Sousa Ribeiro Filho\inst{1,2,3*}\orcidID{0000-0002-8836-4188} \and
 Nikolas Jorge Santiago Carneiro\inst{3}\orcidID{0000-0002-5097-0772} \and
 Lucas Felipe Ferraro Cardoso \inst{2,3}\orcidID{0000-0003-3838-3214} \and
 Ronnie Cley de Oliveira Alves\inst{1,3}\orcidID{0000-0003-4139-0562}
 }

\authorrunning{J. Ribeiro et al.}
%
 \institute{Federal University of Pará (UFPA), Belém, Brazil \and
 Federal Institute of Education, Science and Technology of Pará (IFPA), Ananindeua, Brazil \and
 Vale Institute of Technology (ITV DS), Belém, Brazil\\
 $*$Corresponding Author:\\
 \email{E-mail: jose.ribeiro@ifpa.edu.br}\\
 Phone: +55 91 98185-3166\\
 }

\maketitle              
\begin{abstract}

Strategies based on Explainable Artificial Intelligence (XAI) have promoted better human interpretability of the results of black box models. This opens up the possibility of questioning whether explanations created by XAI methods meet human expectations. The XAI methods being currently used (\textit{Ciu}, \textit{Dalex}, \textit{Eli5}, \textit{Lofo}, \textit{Shap}, and \textit{Skater}) provide various forms of explanations, including global rankings of relevance of features, which allow for an overview of how the model is explained as a result of its inputs and outputs. These methods provide for an increase in the explainability of the model and a greater interpretability grounded on the context of the problem. Intending to shed light on the explanations generated by XAI methods and their interpretations, this research addresses a real-world classification problem related to homicide prediction, already peer-validated, replicated its proposed black box model and used 6 different XAI methods to generate explanations and 6 different human experts. The results were generated through calculations of correlations, comparative analysis and identification of relationships between all ranks of features produced. It was found that even though it is a model that is difficult to explain, 75\% of the expectations of human experts were met, with approximately 48\% agreement between results from XAI methods and human experts. The results allow for answering questions such as: ``\textit{Are the Expectation of Interpretation generated among different human experts similar?}'', ``\textit{Do the different XAI methods generate similar explanations for the proposed problem?}'', ``\textit{Can explanations generated by XAI methods meet human expectation of Interpretations?}'', and ``\textit{Can Explanations and Expectations of Interpretation work together?}''.

\keywords{Explainable Artificial Intelligence  \and Black Box Model \and Human in the Loop \and Homicide prediction \and Machine Learning}
\end{abstract}
\section{Introduction}

In recent years, technology has increasingly evolved and allowed intelligent algorithms to be present in our daily lives through solutions to the most diverse types of problems, thus further requiring that machine learning models solve increasingly complex problems provinding confident explainabilities of their decisions \cite{shalev2014understanding,ghahramani2015probabilistic}.

Computational models based on bagging and boosting algorithms, because they provide high performance and high generalization capacity, are commonly used in computing to solve regression and classification problems based on tabular data. However, these models are not considered transparent algorithms\footnote{Transparent Algorithms: Algorithms that generate explanations for how a given output was produced. Examples: Decision Tree, Logistic Regression and K-nearest Neighbors \cite{arrieta_explainable_2019_20}.}, being considered black box algorithms\footnote{Black Box Algorithms: Machine learning algorithms that have classification or regression decisions that are hidden from the user \cite{cortez2011opening_blckbox_opening}.} and, therefore, are less used in problems related to sensitive contexts, such as health and safety \cite{machine_learning_ML_health,machinelearning_ML_safety}.

By observing the most recent literature on Explainable Artificial Intelligence (XAI) \cite{gunning2019xai}, the use of black box algorithms in sensitive real-world contexts requires confidence (on the part of the human user) to be gained in the predictions of this type of algorithm. In this sense, different strategies have been developed on two knowledge fronts: one aimed at generating greater explanations of the model itself; and other front with analyzes concerning the interpretation of the explanations produced (interpretations made by a human user) \cite{arrieta_explainable_2019_20,interpretabilidade_messalas2019,interpretabilidade_review2018}. 

Black box model explanations are created through analyzes \textit{Model Agnostic}\footnote{\textit{Model Agnostic}: does not depend on the type of machine learning model \cite{molnar2020interpretable}.} or \textit{Model Specific}\footnote{\textit{Model Specific}: depend of the one specific type of machine learning model \cite{khan2022model_specific}.}, also referred to as \textit{Model Inductions} \cite{molnar2020interpretable,darpa_2019} or even \textit{Post-hoc Analyzes} \cite{arrieta_explainable_2019_20}, since in this type of technique only the training data, test data, the model itself and its outputs are used for creating explanations.

The limited understanding of black box models requires the search for methods and tools that can provide information about local explanations --- aiming at predicting around an instance through various methods to obtain a local feature relevance ranking \cite{xai_local_global_2020} ---  and global explanations --- when it is possible to understand the rationale of all instances of the model by generating a global feature relevance ranking \cite{xai_local_global_2020,guidotti2018survey} --- as a means of making interpretable, and thus more reliable, decisions \cite{darpa_2019}.


The term \textit{Ethical AI} \cite{muller2021ten}, has been growing in the area of machine learning in recent years, which shows the concern of the computing community with the development and use of models that are based on responsible and reliable practices in the use of AI. As a result, guidelines, tools and new methods have emerged with the aim of explaining machine learning models, making them more reliable, since a human can only trust what they can understand \cite{angerschmid2022fairness}.

The terminologies Feature Relevance Ranking and Feature Importance Ranking are widely used as synonyms in the computing community, but have different definitions in XAI study area, as shown in \cite{arrieta_explainable_2019_20}. Since feature rankings are regarded as ordered structures whereby each feature of the dataset used by the model appears in a position indicated by a score. The main difference being that, in relevance ranking, the calculation of the score is based on the model output, whereas to calculate the importance ranking of features, the correct label to be predicted is used \cite{arrieta_explainable_2019_20,molnar2020interpretable}.



In previous studies \cite{ribeiro_complexity_et_al_2021}, evidence was verified that shows the existence of models (datasets and algorithms) that are easy to explain and also difficult, through analyzes involving 82 different models (different algorithms and datasets). Since the use of several XAI methods in explaining a single model can allow the generation of different explanations based on relevance ranks --- which show that the model is difficult to explain --- or even similar explanations between the methods --- which show that the models is easy to explain.


Seeking to continue the results found in \cite{ribeiro_complexity_et_al_2021}, this article carries out specific studies from the perspective of just one model, duly evaluated in \cite{ribeiro_pred2town_et_al_2021}, seeking to bring information about the context in which the model is inserted and how these aspects imply in their explanations aiming for greater confidence in the model. 

A technique is also presented, called \textit{ConeXi}, which allows the combination of different explanations coming from XAI methods or even human people (called here Expectation of Interpretation). Enabling the insertion of humans in the explanation process, as in \cite{peterflach_humanintheloop,human_in_the_loop_zanzoto,sokol2020explainability_peterflash_humanintheloop,wilming2022scrutinizing_ML_human_in_the_loop}. 

Given this context and the various research fronts involving explanations, interpretations and human interactions in the black box opening process\footnote{Black box opening: Set of methods, strategies and processes used to make black box models explainable \cite{cortez2011opening_blckbox_opening}}, the following questions arise: 
\begin{itemize}
    \item \textit{``Do the different XAI methods generate similar explanations for the proposed problem?''};
    \item \textit{``Are the expectation of interpretation generated among different human experts similar?''};
    \item \textit{``Can explanations generated by XAI methods meet human expectation of interpretations?''};
    \item \textit{``Can Explanations and Expectations of Interpretation work together?''}.
\end{itemize}

By seeking to answer these questions, an experiment was developed that uses the machine learning model of homicide prediction advocated in \cite{ribeiro_pred2town_et_al_2021}, and from this, the 6 rankings explanations  were generated by means of XAI methods, as in \cite{ribeiro_complexity_et_al_2021}, and 6 ranks expectation of interpretations generated by different human experts. 

Then, comparisons and identification of existing relationships between all pairs of ranks created were performed to find the desired answers. Finally, the generated ranks were combined into a single overall rank by means of a technique proposed hereby based on the results of the explanations of the XAI methods and expectation of interpretations.  

The main contributions of the research to the area of machine learning, which can be completely replicated or used for other research and contexts, are:

\begin{itemize}

    \item Discussion regarding the similarity of explanations generated by XAI methods and their interpretability, focusing on the specific context-sensitive problem --- homicides prediction --- in order to measure whether the XAI methods explain the model as expected by human experts;
    
    \item Concept of Expectation of Interpretation, which in general terms is the interpretation expected by an expert of a real-world problem based on their knowledge of the problem and the working principle of the machine learning model being analyzed;
    
    \item The ConeXi, a tool to combine Expectation of Interpretation with explanations by XAI methods, which are based on global feature relevance rankings, in order to build a Collaborative Explanation of the model using human expert knowledge and different XAI methods, i.e. human and machine.
    
    \item Overall methodology developed by this study as a deliverable, as it promotes data used, code developed, results collected, and the repositories created, in accordance with the \textit{Fair Guiding Principles} for scientific data management and stewardship.
    
\end{itemize}

\section{Background}


This section will present: The concepts of explainability and interpretability in XAI; The operating principles of the XAI methods based in relevance ranks; And aspects referring to previous research on models considered easy and difficult to explain.

\subsection{Explainability and Interpretability in XAI}

The concepts of explainability and interpretability in machine learning are considerably close and even complement one another \cite{arrieta_explainable_2019_20,molnar2020interpretable}. Therefore, it is of utmost importance that they are presented and differentiated.

Explanability is associated with the explanatory interface between a computational model and a human, which aids in the decision-making process as it seeks to make the model understandable \cite{arrieta_explainable_2019_20,molnar2020interpretable}.

Interpretability is the ability to provide meaning in terms that are understandable to a human being, or even the attempt to interpret an explanation \cite{arrieta_explainable_2019_20,interpretabilidade_singh2020model,interpretabilidade_messalas2019,molnar2020interpretable,interpretabilidade_review2018}.

Based on these two concepts, which are widespread in the area of machine learning, it is understood that in a practical way explainability seeks to create subsidies that explain the black box model\footnote{Explain the black box model: Also known as the process of ``opening the black box".} in a technical manner, whereas  interpretability is used-centric wich meaning to the explanations created for a human user, such meaning being based on the context of the problem and the knowledge of the individual \cite{molnar2020interpretable,arrieta_explainable_2019_20}.

Both explainability and interpretability of models are fundamental pieces in the decision-making process, as they provide the end user with support in detecting various problems or even biases in the data being used by the model \cite{arrieta_explainable_2019_20,interpretabilidade_singh2020model}.

It is not possible to conduct a study involving analyzes of explainability and interpretability of computer models without considering the specific context/problem in which they are embedded and the human factors as well \cite{arrieta_explainable_2019_20}.

In this sense, this research focuses on a single specific problem to perform its analyzes. In addition to this and also to issues of time and cost feasibility, the context of homicide prediction was chosen.

Therefore, it can be assumed that explainability and interpretability allow the generation of reliability, understanding, and fairness to black box machine learning models. In the studies and experiments described herein, the main focus is on the explainability of each generated model and its relationship to the interpretabilities (in this case, expectations) generated by humans in the context of crime prediction.

\subsection{Methods of Explainable Artificial Intelligence}


In recent years, there has been an increasing need to explain black box machine learning models in an agnostic and specific manner. Among the various initiatives present in the literature, there is a greater number of XAI methods developed specifically for neural networks, whereas a smaller number of methods are specifically developed for tree-ensemble algorithms \cite{explicabilidade_rede_neural,kindermans_learning_2017_2018,review_xai_2021,ethical_ml_git}.


The need to obtain greater confidence in black box models, currently the community in the XAI area has been developing various methods, concepts, techniques and tools in order to carry out the process of explaining these models. Thus, it is argued that from the creation of layers of explanations on the model, a human user can create their interpretations and thus better understand how the decisions taken by the model were carried out, obtaining greater confidence at the end of the process \cite{arrieta_explainable_2019_20,molnar2020interpretable}.


The so-called \textit{post-hoc} explanation is the currently most widely used existing XAI method category in the computing community. Their main peculiarity is the fact that they only use training data, test data, model output data and the model itself, already duly trained, to generate the explanations\cite{arrieta_explainable_2019_20}.

According to \cite{molnar2020interpretable}, the \textit{post-hoc} XAI techniques can be divided into different strategies: \textit{Text Explanations}, \textit{Visual Explanations}, \textit{Local Explanations}, \textit{Explanation-by-simplification}, \textit{Feature Relevance Explanations} and \textit{Explanation-by-example}. Based on these types of methods, this research focus only on the \textit{Feature Relevance Explanations}, because the ranking structure makes it possible to carry out a quantitative comparative analysis of the explanations generated.

Based on the above, this research conducted a bibliographic and practical survey (development) on the main existing XAI methods, specifically aimed at generating model-agnostic or model-specific global explanation ranks that support tabular data and tree-ensemble algorithm.

As a result, a total of six tools were selected, with libraries updated and compatible with each other, with current Python development platforms and Scikit-Learning algorithms \cite{scikit-learn}. These tools are: \textit{CIU} \cite{ciu_ref}, \textit{Dalex} \cite{dalex_r_ref}, \textit{Eli5} \cite{eli5_ref}, \textit{Lofo} \cite{lofo_ref}, \textit{SHAP} \cite{shap_ref} e \textit{Skater} \cite{skater_ref}.

The methods referenced herein generate explanatory ranks based on the same previously-trained machine learning models, manipulate their inputs or/and produce new intermediate models (copies). Therefore, a comparison of the generated explanatory ranks is fair and feasible.

It should be noted that during the initial analyzes and executions, as occurred in \cite{ribeiro_complexity_et_al_2021}, this research found XAI methods with incompatibilities at the version of libraries and dependencies, making it impossible to use and compare some methods. For this reason, some XAI means were not used for this research (for example: \textit{Alibe-ALE} \cite{alibi_ale_ref}, \textit{Lime} \cite{lime_ref}, \textit{Ethical ExplainableAI} \cite{ethical_xai_site}, \textit{IBM Explainable AI 360} \cite{ibm_xai360}, \textit{Interpreter ML} \cite{interpretML_arxiv}, \textit{Anchor} \cite{ribeiro2018anchors} and \textit{QLattice} \cite{wenninger2022explainable_qlattice}).

A general comparison of the main characteristics of the methods that meet the requirements of this research is shown in table \ref{tab_resumo_xai}.

\begin{table*}[h]
\centering
\caption{Main XAI methods surveyed}
\resizebox{1\textwidth}{!}{%
\begin{tabular}{c|c|c|c|c|c|c}
\hline
\textbf{\begin{tabular}[c]{@{}c@{}}Name\end{tabular}} & 
\textbf{Ref.} &
\textbf{Base algorithm} &
\textbf{\begin{tabular}[c]{@{}c@{}}Explanation\\ technique \end{tabular}} &
\textbf{\begin{tabular}[c]{@{}c@{}}Global \\ explanation\\ (by rank)\end{tabular}} &
\textbf{\begin{tabular}[c]{@{}c@{}}Local \\ explanation\end{tabular}} &
\textbf{\begin{tabular}[c]{@{}c@{}}Model\\ Specific or \\ Agnostic?\end{tabular}} \\ \hline
\textit{CIU} & \cite{ciu_ref} & \begin{tabular}[c]{@{}c@{}}Decision\\Theory \end{tabular} & \begin{tabular}[c]{@{}c@{}} Feature Permutation \\and Multiple Criteria\\Decision Making \end{tabular} & Yes & No & Agnostic \\ \hline
\textit{Dalex} & \cite{dalex_r_ref} & \begin{tabular}[c]{@{}c@{}}Leave-one\\ covariate out\end{tabular} & Feature Permutation & Yes & Yes & Agnostic \\ \hline
\textit{Eli5} & \cite{eli5_ref} & \begin{tabular}[c]{@{}c@{}} Assigning \\weights\\ to decisions \end{tabular} & \begin{tabular}[c]{@{}c@{}}Feature Permutation\\and Mean Decrease\\ Accuracy\end{tabular} & Yes & Yes & Specific \\ \hline
\textit{Lofo} & \cite{lofo_ref} & \begin{tabular}[c]{@{}c@{}}Leave One\\ Feature Out\end{tabular} & Feature Permutation & Yes & No & Specifc\\ \hline
\textit{SHAP} & \cite{tree_shap_ref} & Game Theory & Feature Permutation & Yes & Yes & Specific \\ \hline
\textit{Skater} & \cite{skater_ref} & \begin{tabular}[c]{@{}c@{}}Information\\ Theory\end{tabular} & Feature Relevance & Yes & Yes & Agnostic \\ \hline
\end{tabular}
}
\label{tab_resumo_xai}
\end{table*}

All the tools described above are capable of generating different types of explanations (instance-level explanation, dependent plot, impacts on the model output, similarity of instances, among others), which go beyond global explanations based on feature ranks; therefore, the focus of this article is to compare this most basic unit of model explanation, i.e., the global relevance rank.

Even knowing that each of the XAI methods used in this research are based on different algorithms and methodologies, a comparison of their results is fair. Since it will only compare the basic structure of your explanations based on feature rank.

In order to present the basic operating principle of each XAI method, brief definitions of their operation are presented below.

\subsubsection{\textit{CIU}\\}

The \textit{Contextual Importance and Utility (CIU)} is a XAI method based on Decision Theory \cite{decisions_1993} that focuses on serving as a unified metric of model-agnostic explainability based on the implementation of two different scores: Contextual Importance (CI) and Contextual Utility (CU) \cite{ciu_ref}.

Contextual Importance is linked to the idea of measuring, through different values, how important a feature is in a specific general context of the dataset used as a basis for the model \cite{ciu_git}.

Contextual Utility is linked to the idea of measuring, through different utility values, how much an attribute is used by different smaller contexts present in the data on which the dataset is based \cite{ciu_git}.

As verified in preliminary tests carried out by this research, these two indices generate equal ranks. Thus, it was decided to use the \textit{CI} \cite{ciu_ref}, since its definition shows that this score is more general to the context of the model data \cite{ciu_ref}.

\subsubsection{\textit{Dalex}\\}

The method \textit{Dalex} is a set of XAI tools based on the \textit{LOCO} (\textit{Leave one Covariate Out}) approach and can generate explainabilities from this approach. This method receives the model and the data to be explained, calculates model performance, performs new training processes with new generated data sets, and makes the inversion of each feature of the data in a unitary and iterative way, methods what features are important to the model, evaluates its performance obtained according to the inversions of the features\cite{dalex_python_ref,dalex_book}.

The feature inversions performed by this XAI method are literal when observing a dataset --- that is, the feature has its indexes inverted ---, for example: in a dataset that has the features $F1$ and $F2$ together with their respective values $V_1, V_2, V_3, ..., V_n$ and $V_1, V_2, V_3, ..., V_n$, when inverting the values present in $F_1$ their values are $V_n, V_{n-1}, V_{n-2} , ...,V_1$ \cite{dalex_python_ref}.

\subsubsection{\textit{Lofo}\\}

A little less popular but very powerful is the \textit{Leave One Feature Out (Lofo)}, a XAI method with a similar proposal to that of \textit{Dalex}, but no feature inversion is performed here, because in the \textit{Lofo} metric the iterative step is based on iterative removal of the features to find its global relevance to the model. This method also analyzes the performance of the model \cite{lofo_ref}.

\textit{LOFO} initially evaluates the model's performance with all input features included, then iteratively removes one feature at a time, retrains the model and evaluates its performance on a specific validation dataset. The mean and standard deviation of the relevance of each feature are then reported \cite{lofo_ref}.


\subsubsection{\textit{Eli5}\\}

A very popular and quick method to be performed, the \textit{Explain Like I’m Five (Eli5)} is a complete tool that helps explore machine learning classifiers and explains the predictions \cite{eli5_ref,eli5_git}.

Among the many different ways to explain a machine learning model, \textit{Eli5} is capable of executing the \textit{Mean Decrease Accuracy (MDA)} algorithm in order to generate an attribute relevance ranking \cite{eli5_git}.

The central idea of the algorithm is to calculate the relevance of the feature by observing how much the performance (accuracy, F1, R$^2$, etc.) decreases when a feature is not available for the model. Thus, each feature is removed (only from the test part of the dataset) and then the model's performance is calculated without it using a specific feature \cite{eli5_ref}.

When it is said to remove the feature, it must be understood that its values are replaced by random noise, thus not containing useful information for the model. In this method, the noise is extracted from the same distribution as the original feature values (otherwise, the estimator may fail). And so the permutation relevance is calculated in Eli5 \cite{eli5_ref}.

\subsubsection{\textit{Shap}\\}

One of the most popular and currently used methods is the \textit{SHapley Additive exPlanations (SHAP)}, proposed as a unified method of feature relevance that explains the prediction of an instance $X$ from the contribution of an feature. What makes this method different from the others is the calculation explanation score based in the game theory of \textit{Shapley Value} \cite{roth1988shapley,tree_shap_ref,shap_ref}.

For \textit{Shap}, each feature is considered a player in a game and the objective of this game is to achieve the model's output, that is, the prediction itself. In this way, in an iterative manner on feature subsets, new models are re-trained with and without the feature subsets and the \textit{Shapley Values} of each feature are calculated, ultimately generating an attribute relevance ranking \cite{shap_ref,shap_doc}.

This research chose to use the version of \textit{Shap} that is specific for decision tree-based models, due to a lower computational cost than the agnostic version.

\subsubsection{\textit{Skater}\\}

Last but not least is the method \textit{Skater}, a set of tools capable of generating ranks of the relevance of model features, differing from the other methods to calculate its explanation index based on Information Theory \cite{info_theory_1994}, through measurements of entropy in changing predictions through a disturbance of a certain feature \cite{skater_ref}.

Unlike other XAI methods, Skater was developed by a private initiative and is now closed source software. In-depth information about how this method works is very scarce, but it is still widely used by the XAI community \cite{skater_ref,skater_git}.

What is known is that the global explanations of the model, Skater makes use of feature relevance independently of the model used and makes use of \textit{Partial Dependency Plots (PDP)} to judge the bias of a model and understand its general behavior \cite{skater_ref,skater_citation}.

\subsection{Difficulty of Explaining a Black Box Model}

The complexity of a model can be attributed to the algorithm used in the prediction process and also to the properties of the dataset that this model seeks to generalize. Such properties may or may not have a direct relationship with the context of the problem, as they may simply be technical alternatives for processing the features used in the model inputs \cite{ribeiro_complexity_et_al_2021}.

In the study \cite{ribeiro_complexity_et_al_2021}, which precedes the research described in this article, comparative analyzes were carried out between 41 different tree-ensemble machine learning models (based on \textit{Random Forest}, Gradient \textit{Boosting} algorithms and 41 different datasets referring to classification problems binary), aiming to identify the correlations between the relevance ranks of attributes generated from 6 XAI methods (\textit{Ciu, Dalex, Eli5, Lofo, Shap,} and \textit{Skater}) applied to each model created.

In \cite{ribeiro_complexity_et_al_2021}, the two algorithms were used to create two machine learning models, for each dataset, totaling 82 models created. These models are properly: trained with training and test splits proportional to 70\% and 30\% respectively, passed through the tuning process, grouped using a clustering algorithm and finally explained by XAI methods.

Finally, \textit{Spearman} correlations were calculated for all pairs of feature relevance ranks, allowing the identification of evidence about the existence of tree-ensemble models that are easier to explain and also more difficult.

These two important findings are based on logical reasoning that, without taking into account the context in which the model is inserted, can analyze it as difficult or easy to explain. To achieve this, it is understood that by having a tree-ensemble model to be explained through various XAI methods:

\begin{itemize}
    \item Similar explanations can be obtained between the various methods, indicating that this analyzed model can be considered easier to explain, since there is an explanatory concordance for its variables even using different XAI methods. Therefore, it is easier for a human to interpret or trust their explanations.

    \item Divergent explanations can be obtained between the various methods, indicating that this model can be considered more difficult to explain, because if each method explains the same model using different feature relevance rankings, it becomes difficult for a human to interpret or trust the explanations of model.
    
\end{itemize}
  
The findings of the study \cite{ribeiro_complexity_et_al_2021} are based on a high number of models and datasets with different complexity, but even given these results, it is necessary to explore in greater detail the evidence found leading the specific context in which the model is inserted is taken into account.


\section{Related works}

This work is based on current literature related to Explainable Artificial Intelligence, in the following sections research works related to the two main points involving the problem of this research will be presented and discussed: Machine Learning in Criminology and XAI Method Analyzes.

\subsection{Machine Learning in Criminology}
Criminology is the field of study that brings together knowledge related to crimes associated with society and their causes \cite{teoria_crime_1968}. Thus, criminological data are documentary reflections on crimes that occurred in the real world \cite{teoria_crime_cano_as_2002}.

Typically, criminological data is not easily accessible, as it often represents information from confidential, sensitive and complex contexts related to crimes of different natures \cite{teoria_crime_1968}. Even with these peculiarities, in recent years the emergence of research in the area of machine learning involving criminological data has been observed, as these models have helped in understanding how crime manifests itself \cite{al2019_crime_1,ang2015san_crime_2,damasceno2011prediction_crime_3,jin2020addressing_crime_4,shermila2018_crime_5,yadav2017_crime_7,alderden2007_crime_8,colasanti2023homicide_crime_9}, which has the following main characteristics:

The research \textit{``Crime Type Prediction''} \cite{al2019_crime_1}, shows how criminology data can be used together with climate data on a temporal basis in order to make it possible to predict the occurrence of specific types of crimes. For this, algorithms \textit{Logistic Regression, Random Forest, Decision tree,} and \textit{XGBoost} were used and historical climate data between the years 2012 to 2017, all referring to the city of \textit{Chicago} (\textit{United States}). Similar research to that found in \cite{alderden2007_crime_8}.

Following an idea similar to that of the research above, the work \textit{``San Francisco Crime Classification''} \cite{ang2015san_crime_2} is cited, which, through machine learning models based on the algorithms \textit{Logistic Regression, Naive Bayes,} and \textit{Random Forest} and criminological data from the city of \textit{San Francisco} (\textit{United States}) from 2003 to 2015, can predict the type of crime that will occur. Other research initiatives with data from the same city can be seen at \cite{jin2020addressing_crime_4,shermila2018_crime_5}.

As an example of research that uses exclusively transparent algorithms together with criminology data, the research \textit{``Crime Pattern Detection, Analysis \& Prediction''} \cite{yadav2017_crime_7} is presented, which uses the algorithms \textit{K-means}, \textit{Naive Bayes}, \textit{Apriori}, and \textit{Association Rule}, in the process of extracting associations in crime data from the city of \textit{Thane} (\textit{India}). It can be seen here that the most important thing is not prediction, but rather recognition of associative patterns related to base crimes.

Leaving classic machine learning algorithms and moving towards algorithms aimed at predicting time series, we have the research \textit{``Homicide-Suicide in Italy Between 2009-2018: An Epidemiological Update and Time Series Analysis''} \cite {colasanti2023homicide_crime_9}, which uses \textit{ARIMA} in the process of predicting homicide-suicide crimes, using data from the city of \textit{Rome} (\textit{Italy}), covering the years 2009 to 2018.

The occurrence of crimes is directly linked to complex social relationships that human beings build while living in society \cite{cerqueira_crimes_scielo}, so socioeconomic data can be strong allies in the process of understanding the motivations for committing a crime. In this sense, there is the research \textit{``A Prediction Model for Criminal Levels Specialized in Brazilian Cities''} \cite{damasceno2011prediction_crime_3}, which presents a proposal for a model based on a Neural Network aimed at predicting crime levels, in the city of \textit{Fortaleza} (\textit{Brazil}), using historical crime and socioeconomic data between the years 2007 and 2008.

The research entitled \textit{``Prediction of Homicides in Urban Centers: A Machine Learning Approach''} \cite{ribeiro_complexity_et_al_2021} introduced the research defended here, and it presents a rigorous process of selecting a better model ( among a total of 11 algorithms) capable of carrying out the process of predicting homicide crimes with generic data from the city of \textit{Belém} (\textit{Brazil}).

As above, it can be seen that in experiments involving machine learning models, both transparent algorithms and black box algorithms were used, due to the need to have explanations of how predictions are made, at the same time as there is a need to have high performance, even if they are different models \cite{al2019_crime_1,ang2015san_crime_2,shermila2018_crime_5,yadav2017_crime_7,alderden2007_crime_8}.

Therefore, using Explainable Artificial Intelligence techniques in a problem that involves criminological data is to meet an emerging need in the area of Criminology, bringing important contributions to society. Providing the generation of black box model explanations that generally present the highest performance on these types of data.

All the works listed above are directly related to the crime prediction model used in this study, as they served as inspiration and base for the way in which this model uses and generates input and output data respectively.

\subsection{XAI Method Analyzes}

The area of XAI has been experiencing rapid growth, resulting in notable taxonomic differences in the terms used by different works. Seeking to standardize the understanding of the taxonomic terms used here, it was decided to adopt the taxonomy used in the work \cite{arrieta_explainable_2019_20} as the basis of this research.

The book that defines the different types of XAI methods that generate model-agnostic global explanations is \cite{molnar2020interpretable}, which presents the following division of types \textit{Partial Dependence Plot (PDP), Accumulated Local Effects (ALE), Feature Interaction, Global Surrogate, Prototypes and Criticisms,} and \textit{Permutation Feature Relevance (Importance)}, the latter being the focus of the research described here.

The article \textit{``Does Dataset Complexity Matters for Model Explainers?''} described by \cite{ribeiro_complexity_et_al_2021}, brings a relevant discussion about the existence of models that are easier to explain and models that are more difficult to explain, using a set of 41 datasets and the \textit{Random Forest} and \textit{Gradient Boosting} algorithms, evidence was verified that the generated models --- composed of the dataset together with the base algorithm --- could present higher correlations or lower correlations among the explanations of the XAI methods \textit{Ciu, Dalex, Eli5, Lofo, Shap,} and \textit{Skater} generated for each model.

Comparing explanations of XAI methods is a recent type of analysis in XAI, as most XAI methods emerged a few years ago and are still beginning to be used by the computing community \cite{holzinger2020explainable}. Therefore, to date there are few works that seek to discuss these comparisons or even use more than a single method to explain models, such as \cite{sahatova2022overview_comparacao_shap_lime,chadaga2023artificial_compare,hariharan2023xai_compare,jouis2021_anchors}. In all the research cited, real-world problems are used as case studies.


For example, in \cite{sahatova2022overview_comparacao_shap_lime} visual comparisons are made of the local explanations produced by the methods \textit{Shap} and \textit{Lime} under a machine learning model aimed at detecting objects in images of tomography. The main comparisons made between the methods are visual, as the model input data is in image format.


In the study \cite{chadaga2023artificial_compare}, a decision support system is created based on the methods \textit{Shap}, \textit{Eli5}, \textit{QLattice}, \textit{Anchor}, and \textit{ Lime} with the aim of creating explanations of a machine learning model aimed at diagnosing \textit{COVID 2019}. Here different types (formats) of explanations are generated and are used to discuss different perspectives of the proposed problem.


In the research \cite{hariharan2023xai_compare} the methods of \textit{Permutation Importance}, \textit{Shap}, \textit{Lime}, and \textit{Ciu} are used under a machine learning model aimed at the problem of Intrusion Detection focused on cybersecurity. In this study, comparisons of explanations are made using accuracy, consistency and stability.

In \cite{jouis2021_anchors} the \textit{Anchors} and \textit{Attention} methods are used under a job offer prediction model. In the study, different types (formats) of explanations are used, duly compared in a quantitative and qualitative way.

The field of XAI presents many challenges and opportunities, as can be seen in \cite{arrieta_explainable_2019_20}. One of the most relevant challenges is the search for universal explanations of models, but this is a significantly complex challenge, because as can be seen in works above and such as \cite{peterflach_humanintheloop} one explanation does not fit all, i.e. it cannot be expected that a single explanation will be able to meet the needs of everyone involved directly or indirectly in the creation of the model (experts, developers, context of the problem, among others). Thus, there is no baseline that provides explanations that can be assumed to be universally correct --- However, there is a human expert.

Increasingly, the design of XAI methods has moved closer to the human being, in terms of allowing them to take part in the process of creating explanations for models that support decision-making in real-world situations \cite{human_in_the_loop_zanzoto,wilming2022scrutinizing_ML_human_in_the_loop,sokol2020explainability_peterflash_humanintheloop}, this shows a tendency for methods to work collaboratively with humans.


\section{Materials and Methods}

In this section, methodological aspects of the research are presented: Methodology overview; The general characteristic description of the homicide crime prediction model; The definition of ranks generated by XAI methods; The definition of rankings generated by human experts; Analysis of the generated rankings.

\subsection{Methodology overview}
\begin{figure}
\begin{center}
\includegraphics[scale=0.415]{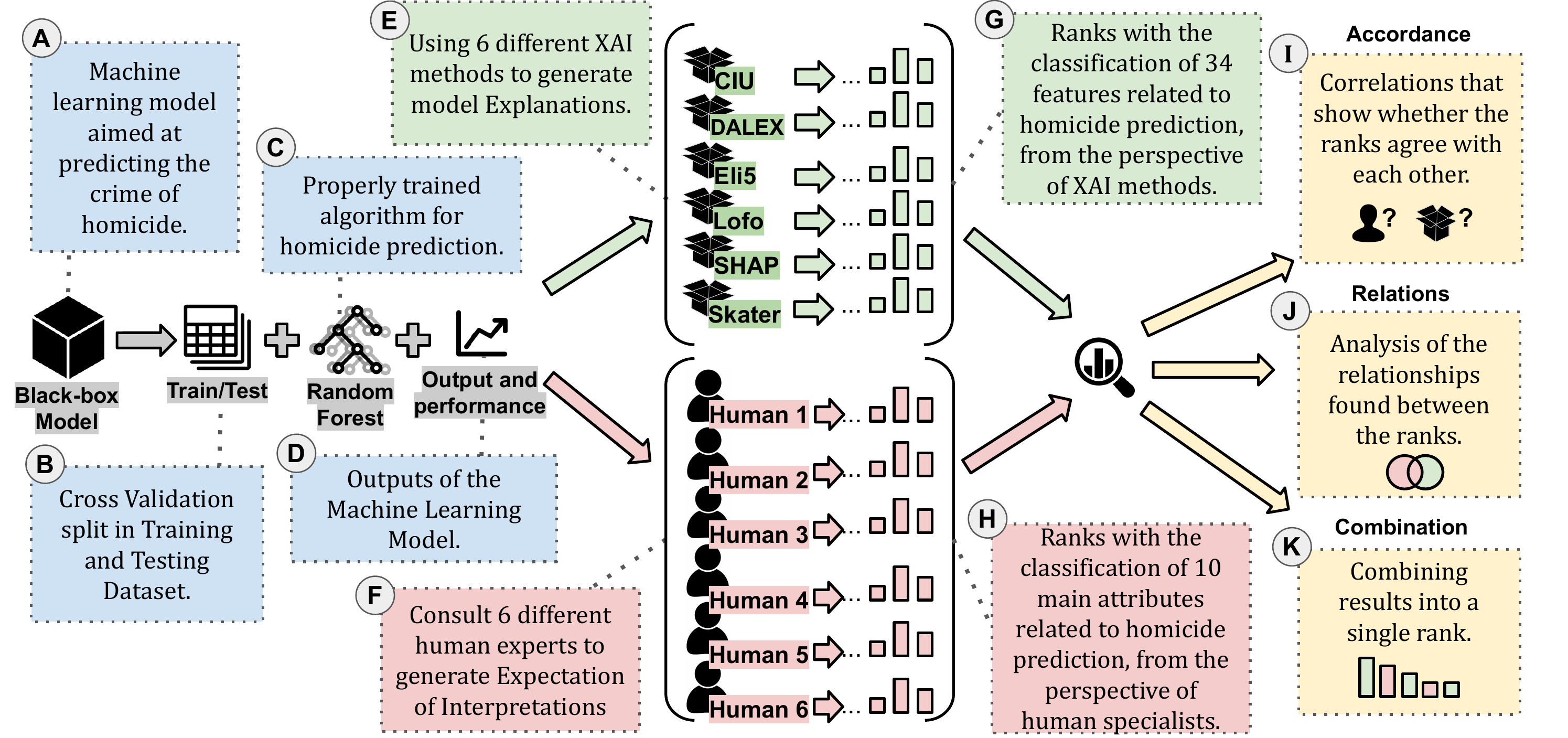}
\caption{Visual scheme of all steps and processes performed by the proposed methodology.}
\label{fig_metodologia}
\end{center}
\end{figure}

A macro view of the main methodological points can be seen in the Figure \ref{fig_metodologia}, which is presented in order to facilitate the abstraction of the steps taken.

This research performed analyzes involving the machine learning model (Figure \ref{fig_metodologia} (A)) as advocated in \cite{ribeiro_pred2town_et_al_2021}, duly trained (Figure \ref{fig_metodologia} (B), (C) and (D)), together with executions of the XAI methods (Figure \ref{fig_metodologia} (E)) and querries with human experts (Figure \ref{fig_metodologia} (F)), so as to generate the explanation ranks (Figure \ref{fig_metodologia} (G)) and expectation of interpretations (Figure \ref{fig_metodologia} (H)), thus allowing for performing analyzes from three different perspectives (Figure \ref{fig_metodologia} (I), (J) and (K)). All the steps in the processes involved in the developed methodology are presented in the following topics.   

All information regarding the reproducibility of the methodology described here can be accessed through the link in the ``Data Available'' area.

\subsection{Prediction of homicides model}

The machine learning model used in the analyzes performed herein has already been validated in \cite{ribeiro_pred2town_et_al_2021}, being developed on the basis of the data provided by the \textit{Secretariat of Intelligence and Criminal Analyzes (SIAC)} of Pará State, Brazil. Such data refer to the police reports registered during the years 2016 to 2018 in the city of Belém, Pará State.

The City of Belém, capital of the state of Pará, Brazil, the scenario of this study, is a city with 1.303.389 inhabitants, according to the last census performed in 2022, and has a human development index of 0.746 \cite{ibge_belem}. This city has institutions that carry out several programs in the area of public security, and in recent years there has been a decrease in cases of violent crimes.

Nevertheless, Belém featured the third worst homicide rate among all Brazilian capitals (74.3) in a 2017 study \cite{ipea_mapa_violencia}. Thus, the data from the police records in this city prove to be interesting sources of information for the herein study.

In a recent survey, Pará State ranks together with eight other Brazilian states as having incomplete data regarding homicide cases and solution thereof \cite{sou_da_paz}. Thus, the importance of the study conducted in \cite{ribeiro_pred2town_et_al_2021} is hereby emphasized, since the developed model proved capable of carrying out the crime prediction process, even under different circumstances. In the future, the introduced methodology may be able to help solve problems related to missing data on violent crimes.

As advocated in \cite{ribeiro_pred2town_et_al_2021}, the machine learning model proposed hereby is based on the dynamics of how different crimes that occur in a city can be explained by different theories in the area of Social Sciences, such as those related to a single person, called ``Theory of Understanding the Motivations of Individual Behavior", and also from the ``Theory of Associated Epidemiology", related to the study of how criminal behavior is distributed and displaced in time and space of a given location \cite{teoria_crime_1968,teoria_crime_cano_as_2002}. 

This second theory, presented above, supports the way in which this research understands the relationships between different types of crimes, that is, through a deep analyzes of the relationships of the most different crimes and their time and space, the possibility of predicting a specific crime is suggested\cite{ribeiro_pred2town_et_al_2021}.

Thus, for the construction of the model, several transformations were performed on the original database of police reports, in order to allow the quantity of each crime occurring in years, months and neighborhoods to be correlated to the binary class referring to the occurrence or not of homicides in the following month, thus making it possible to analyze how different crimes are dynamically related to homicides in the city of study, since the occurrence of specific crimes is related to a context of the conflict between individuals, and one offense may influence homicides \cite{teoria_contra_pessoas_gilaberte2013crimes,teoria_crime_1968,cerqueira_crimes_scielo}.

\begin{center}
\begin{figure}
\includegraphics[scale=0.47]{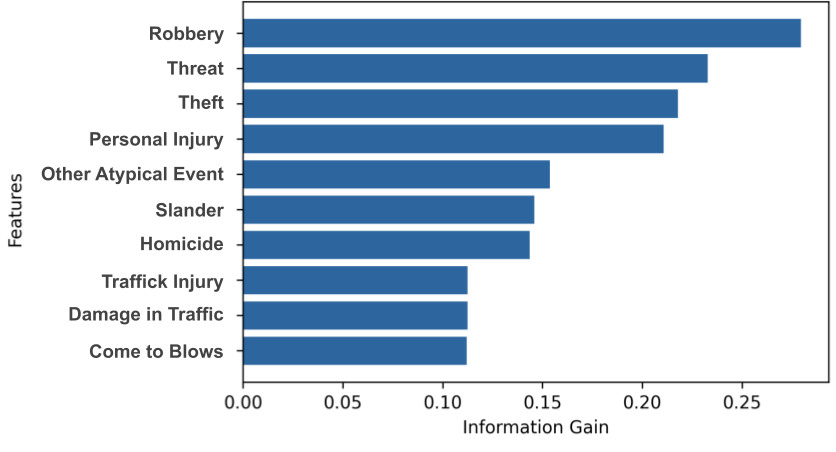}
\captionsource{Rank of the 10 features with the highest values of information gain in the dataset of the crime prediction model.}{\cite{ribeiro_pred2town_et_al_2021}.}
\label{fig_infogain}
\end{figure}
\end{center}

The data used as inputs for machine learning models are completely generic, since they are made up of the month variable (1 variable) along with the different numbers of crimes that occurred in neighborhoods in a specific month (34 variables) and 2.004 instances, which makes this methodology of using criminology data for the prediction of homicides easily replicable to other cities that have the same data. The Figure \ref{fig_infogain} shows the information gain (top 10 features) calculated for the data of the proposed system.

It is also noteworthy that all records of different types of crimes (34 types in total) were selected by a public security specialist. By that, only crime types that can be directly or indirectly linked to homicides are found in the final dataset \cite{ribeiro_pred2town_et_al_2021}. 

The model selection followed the same steps carried out in the research \cite{ribeiro_pred2town_et_al_2021}, aiming to identify the best model (with the best performance) based on the 11 different algorithms: \textit{K-Nearest Neighbors (KNN)} \cite{knn}, \textit{Support Vector Machine (SVM)} \cite{svm}, \textit{Decision Tree (DT)} \cite{dt}, \textit{Neural Network (NN)} \cite{nn}, \textit{Naive Bayes (NB)} \cite{nb}, \textit{Logistic Regression (LR)} \cite{lr}, \textit{Gradient Boosting (GB)} \cite{gb}, \textit{Random Forest (RF)} \cite{breiman2001random_ML_randomforest}, \textit{Extreme Gradient Boosting (XGB)} \cite{xgb}, \textit{Light Gradient Boosting Machine (LGBM)} \cite{lgb} and \textit{CatBoosting (CB)} \cite{cb}.

Firstly, for each of the 11 algorithms, the hyperparameter process was carried out using the \textit{Grid Search with Cross Validation} and stratification of the objective class, in order to optimize the performance of each model. The Table \ref{tab_tunning} shows all the value ranges used in the model properties when carrying out the tuning process, along with the best configurations found.

\begin{table}[]
\begin{center}
\resizebox{.9\textwidth}{!}{%
\begin{tabular}{c|l|l}
\hline
\textbf{Model} & \multicolumn{1}{c|}{\textbf{Range of parameters}}       & \multicolumn{1}{c}{\textbf{Best parameters found}} \\ \hline
               & Learning rate: {[}0.1, 0.5{]};                          & Learning rate: 0.1;                                \\
               & Depth: {[}1, 6, 12{]};                                  & Depth: 12;                                         \\
CB             & Iterations: {[}10, 100, 200{]};                         & Iterations: 200;                                   \\
               & Grow policy: {[}SymmetricTree, Depthwise, Lossguide{]}; & Grow policy: SymmetricTree;                        \\
               & Bagging temperature: {[}0, 0.5, 1{]};                   & Bagging Temperature: 0;                            \\ \hline
               & Min samples leaf : {[}1, 10, 20, 40{]};                 & Min samples leaf : 40;                             \\
               & Max depth: {[}1, 6, 12{]};                              & Max depth: 12;                                     \\
DT             & Criterion: {[}gini,entropy{]};                          & Criterion: entropy;                                \\
               & Splitter: {[}best, random{]};                           & Splitter: random;                                  \\
               & Min samples split: {[}2, 5, 15, 20, 30{]};              & Min samples split: 2;                              \\ \hline
               & Max d.: {[}1, 6, 12{]};                                 & Max depth: 12;                                     \\
               & N. estimators: {[}10, 100, 200{]};                      & N. estimators: 100;                                \\
               & Min samples leaf: {[}1, 10, 20, 40{]};                  & Min samples leaf: 40;                              \\
GB             & Learning r.: {[}0.1, 0.5{]};                            & Learning rate: 0.1;                                \\
               & Loss: {[}deviance, exponential{]};                      & Loss: exponential;                                 \\
               & Criterion: {[}friedman mse, mse, mae{]};                & Criterion: mae;                                    \\
               & Max f.: {[}sqrt, log2{]};                               & Max f.: sqrt;                                      \\ \hline
               & Leaf size:{[}1, 10, 20, 40{]};                          & Leaf size: 1;                                      \\
KNN            & Algorithm:{[}ball tree, kd tree, brute{]};              & Algorithm: ball tree;                              \\
               & Metric: {[}str, callable,minkowski{]};                  & Metric: minkowski;                                 \\
               & N. neighbors:{[}2,4,6,8,10,12,14,16{]};                 & N. neighbors: 8;                                   \\ \hline
               & Learning r.: {[}0.1, 0.5{]};                            & Learning rate: 0.1;                                \\
               & Max d.: {[}1, 6, 12{]};                                 & Max depth: 1;                                      \\
LGBM           & Bootstrap: {[}True, False{]};                           & Bootstrap: True;                                   \\
               & N. estimators: {[}10, 100, 200{]};                      & N. estimators: 100;                                \\
               & Min data in leaf: {[}1, 10, 20, 40{]};                  & Min data in leaf: 40;                              \\
               & Boosting t.: {[}gbdt,dart,goss,rf{]};                   & Boosting type: goss;                               \\ \hline
               & Solver: {[}newton-cg, lbfgs, liblinear, sag, saga{]};   & Solver: sag;                                       \\
LR             & Penalty: {[}l1,l2{]};                                   & Penalty: l2;                                       \\
               & C:{[}0.001,0.008,0.05,0.09,0.1{]};                      & C: 0.1;                                            \\
               & Max iter.: {[}50, 200, 400, 50, 600{]};                 & Max iter.: 50;                                     \\ \hline
NB             & Var. smoothing: {[}1e-5, 1e-7, 1e-9, 1e-10,1e-12{]};    & Var. smoothing 1e-5;                               \\ \hline
               & Learning r.: {[}constant, invscaling, adaptive{]};      & Learning rate: invscaling;                         \\
               & Solver: {[}lbfgs, sgd, adam{]};                         & Solver: adam;                                      \\
NN             & Activation: {[}identity, logistic, tanh, relu{]};       & Activation: tanh;                                  \\
               & Max iter.: {[}200,300,400{]};                           & Max iter.: 300;                                    \\
               & Alpha: {[}0.0001,0.0003{]};                             & Alpha: 0.0001;                                     \\
               & Hidden layer sizes: {[}1,2,3,4,5{]};                    & Hidden layer sizes: 3;                             \\ \hline
               & Max depth: {[}1, 6, 12{]};                              & Max depth: 12;                                     \\
               & Bootstrap: {[}True, False{]};                           & Bootstrap: True;                                   \\
               & N. estimators: {[}10,100, 200{]};                       & N. estimators: 100;                                \\
RF             & Min samples leaf: {[}1, 10, 20, 40{]};                  & Min samples leaf: 1;                               \\
               & CCP alpha: {[}0.0, 0.4{]};                              & CCP alpha: 0.0;                                    \\
               & Criterion: {[}gini, entropy{]};                         & Criterion: gini;                                   \\
               & Max features: {[}sqrt, log2{]};                         & Max feat.: log2;                                   \\ \hline
               & C: {[}0.001, 0.01, 0.1, 1, 10{]};                       & C: 10;                                             \\
SVM            & Kernel: {[}linear, poly, sigmoid{]};                    & Kernel: poly;                                      \\
               & Shrinking: {[}True, False{]};                           & Shrinking: True;                                   \\
               & Degree: {[}1,2,3,4,5{]};                                & Degree: 1;                                         \\ \hline
               & Max d.: {[}1, 6, 12{]};                                 & Max depth: 1;                                      \\
               & N. estimators: {[}10, 100, 200{]};                      & N. estimators: 200;                                \\
XGB            & Min s. le.: {[}1, 10, 20, 40{]};                        & Min samples leaf: 1;                               \\
               & Booster: {[}gbtree, gblinear, dart{]};                  & Booster: gbtree;                                   \\
               & Sampling m.: {[}uniform, gradient based{]};             & Sampling method: uniform;                          \\
               & Tree m.: {[}exact,approx,hist{]};                       & Tree method: approx;                               \\ \hline
\end{tabular}}
\captionsource{All the best parameters found from the execution of the stratified Grid Search with Cross Validation process for the 11 algorithms, based in \textit{AUC} evaluation.}{\cite{ribeiro_pred2town_et_al_2021}.}
\label{tab_tunning}
\end{center}
\end{table}

The \textit{Area Under the Curve (AUC)} was used to evaluation because it takes into account the successes and errors identified in both classes (1 and 0) of the problem in question. Thus, the \textit{AUC} measures both successes and errors of homicides and non-homicides that occurred, a characteristic that is desirable given the nature of the problem — Since predicting a homicide is just as important as predicting a non-homicide.

Next, a performance analysis of the 11 models was carried out, duly tuned by the tuning process, using the \textit{Accuracy (ACC)} measure, obtaining the result shown in the table \ref{tab_accuracy}.

\begin{table}[h!]
\begin{center}
\begin{tabular}{cc|cc|cc}
\hline
\textbf{Model} & \textbf{\textit{ACC}}  & \textbf{Model} & \textbf{\textit{ACC}}  & \textbf{Model}    & \textbf{\textit{ACC}}      \\ \hline
\textit{RF}    & 0.76 & \textit{LR}    & 0.74 & \textit{DT}         & 0.71     \\ \hline
\textit{LGBM}  & 0.75 & \textit{SVM}   & 0.74 & \textit{NB}         & 0.69     \\ \hline
\textit{XGB}   & 0.75 & \textit{CB}    & 0.74 & \textit{KNN}        & 0.69     \\ \hline
\textit{NN}    & 0.74 & \textit{GB}    & 0.72 & \multicolumn{2}{c}{-} \\ \hline
\end{tabular}
\caption{Accuracy by Model. \textbf{Source:}\cite{ribeiro_pred2town_et_al_2021}.}
\label{tab_accuracy}
\end{center}
\end{table}


Based on the results of the accuracy calculated for the 11 models, it is possible to indicate the \textit{RF, LGBM} and \textit{XGB} models with similar performances, being the models that best achieved the two objective classes. However, more in-depth analyzes still need to be carried out, such as: \textit{Stability} and \textit{Statistics}.

\begin{figure}[b!]
\begin{center}
\includegraphics[scale=0.35]{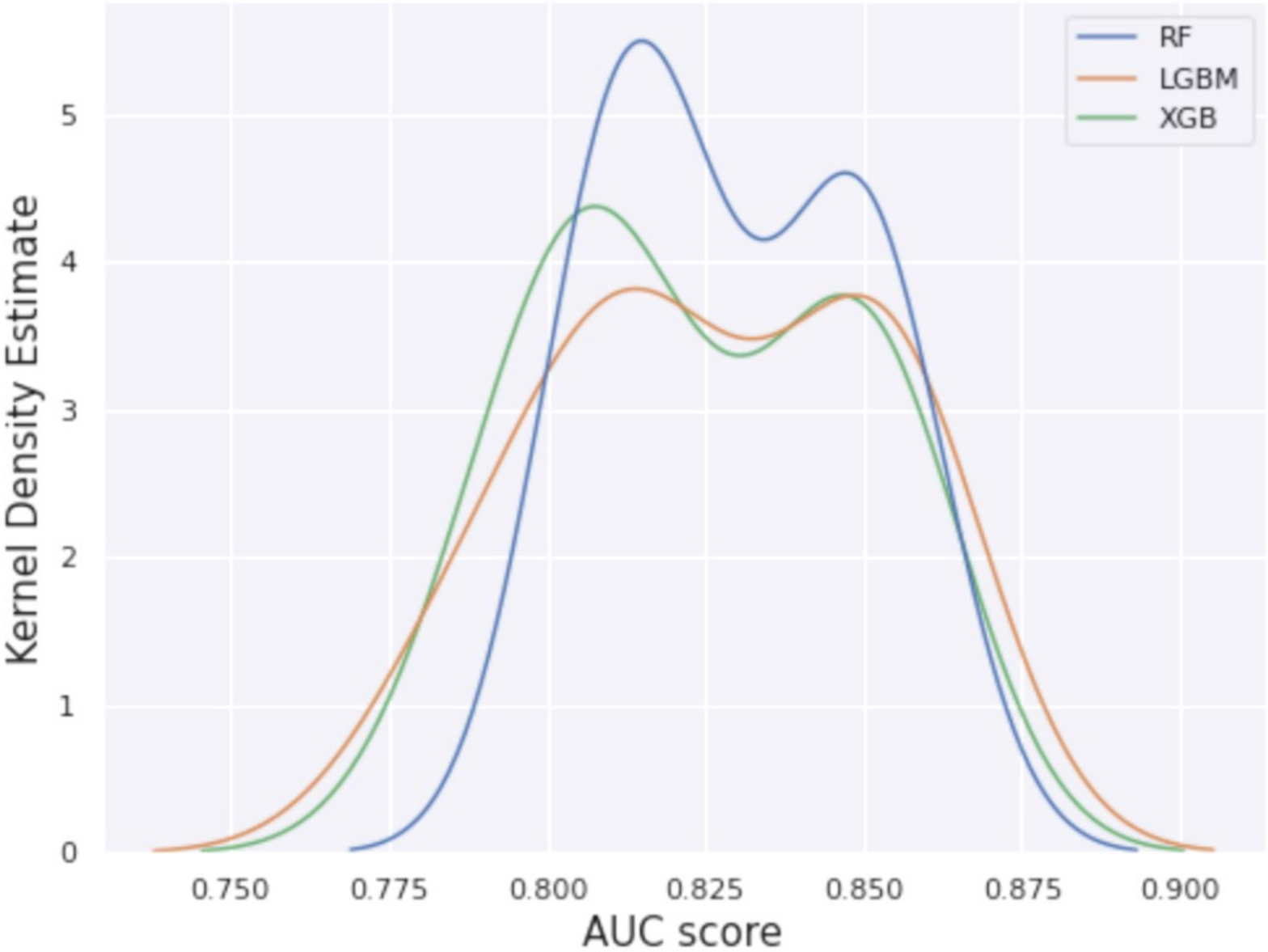}
\captionsource{Density of AUC scores for execution of cross-validation with fold’s size equal to 7 for \textit{RF, LGBM}, and \textit{XGB}.}{\cite{ribeiro_pred2town_et_al_2021}.}
\label{fig_stabilit}
\end{center}
\end{figure}

To evaluate the stability of the models, \textit{Stratified Cross Validation} runs with fold = 7 were performed for each model. After this execution, each model has its outputs analyzed by the \textit{AUC} metric and, finally, a \textit{Kernel Density Estimate} graph (\textit{Gaussian} with bandwidth $0.6$) was created with the executions, aiming to identify the \textit{AUC} value for each one of the 7 folds. This analysis is shown in Figure \ref{fig_stabilit}.

Analyzing Figure \ref{fig_stabilit}, it can be seen that the three algorithms \textit{(RF, LGBM, and XGB)} present similar stabilities, as for each 7 executions, a similar variation in algorithm performance was identified between the models, with values between 0.775 and 0.875 \textit{AUC}.

Even though small, the \textit{RF} algorithm presented greater stability when compared to \textit{LGBM} and \textit{XGB}, since \textit{RF} presented slightly higher density (higher values on the y axis) and concentrated (x axis range).

However, considering the context of the homicide prediction problem, which presents significant sensitivity and the need for an accurate model (high performance). This study recognizes the results of the \textit{Random Forest} algorithm as the best presented for the context in question.

\begin{figure}[!h]
\begin{center}
\includegraphics[scale=0.44]{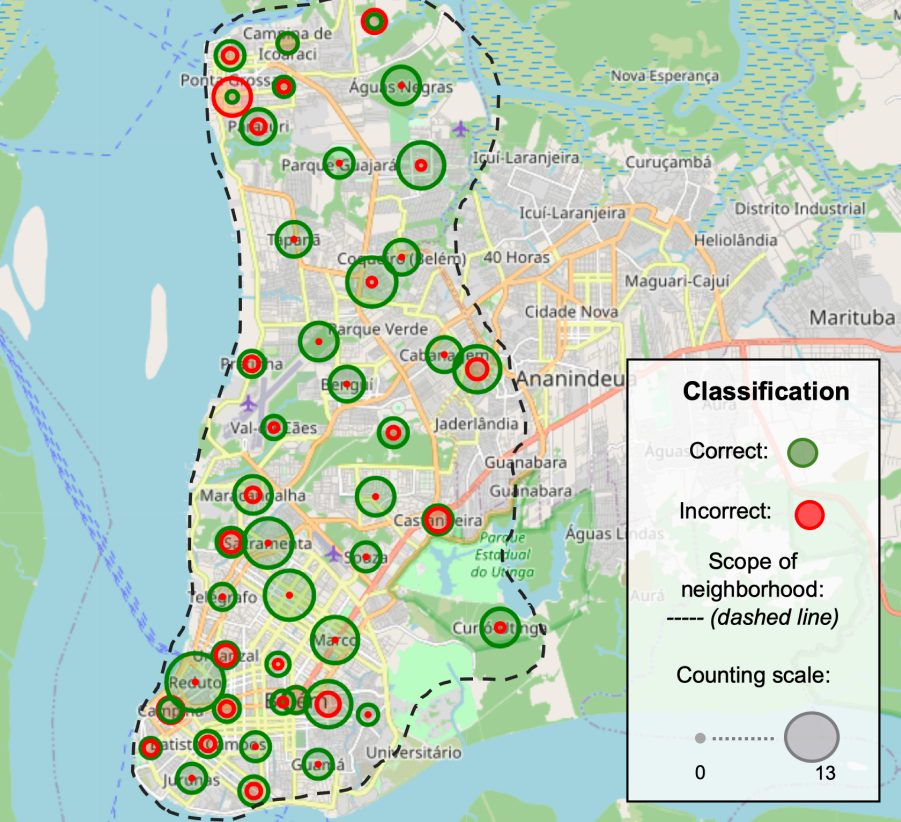}
\captionsource{Distribution of errors and successes of the RF model across neighborhoods in the study city.}{\cite{ribeiro_pred2town_et_al_2021}.}
\label{fig_map}
\end{center}
\end{figure}

An interesting way to observe the model's outputs is by using the map of the study city, as a way of visualizing the spread of the \textit{RF} model's errors depending on the different neighborhoods (metadata) existing in the dataset, Figure \ref{fig_map}.

It can be seen in Figure \ref{fig_map} that the biggest model errors (red circles) are distributed in specific neighborhoods. This may occur due to problems in collecting data from crime reports in specific police stations.

However, in Figure \ref{fig_map} it is possible to visualize the predominance of correct output from the model (green circles) distributed across the map in general.

Thus, considering all the steps carried out in \cite{ribeiro_pred2town_et_al_2021}, shown above, the research described here also assumes the \textit{Random Forest} model as the best machine learning model aimed at the process of predicting homicide crimes (among the models tested) and will therefore use this model aiming to generate the results that will be used later.

More details about the dataset and analyzes of the prediction of homicides model can be found in section \textit{``Availability of data and material''} item A.2.

\subsection{Rank by XAI methods}

As seen earlier, it is known to this research that the XAI methods used (\textit{Ciu, Dalex, Eli5, Lofo, Shap} and \textit{Skater}) have different base algorithms that act in creating the explanations of black box models. However, it is emphasized that all these methods are compatible with each other and are comparable, since they all generate the basic rank structure as an explanation for an analyzed model.

The XAI averages can create ranks of explanations based on the relevance that a particular feature has for the model as a whole. In general, these ranks are ordered based on a \textit{score}, defined by the method and calculated for each feature, thus enabling the generation of ranks for a maximum limit of features \textit{n} unknown to this research \cite{molnar2020interpretable}.

A sensitive, but of utmost importance, detail should be observed and properly described regarding the generation of the ranks, since an XAI method can generate a rank of explainability with more than one feature with the same score. By that, when generating each of the XAI method ranks, each feature with the same score was ordered by its label (feature name), thus ensuring a consistent comparison between the methods.

As above, the 6 XAI methods were executed for the homicide prediction model, and thus 6 explicability ranks were created containing the order of importance of the 35 features in the model.

\subsection{Rank by Human Specialists}

This research was supported by the \textit{State Secretariat for Public Security and Social Defense of Pará (SEGUP)}, which was responsible for providing the necessary data and indicating qualified professionals to contribute to the analyzes carried out.

Through \textit{SEGUP}, this research contacted the people in charge of the \textit{Office of Intelligence and Criminal Analyzes (SIAC)} of Pará State to querry a total of 6 experts in the field of criminology on variables present in the dataset that would be strongly related to homicide crimes. The answer was the assignment of 6 specialists in the field, with different profiles:

\begin{itemize}
    \item \textit{Military Police Officers}: Professionals in the area of public security working in the city of study, experience in external (field) and administrative (office) activities related to security;
    

    \item \textit{SIAC} \textit{and} \textit{SEGUP} \textit{Management Servers}: Public security professionals working in the study city, considerable experience in activities carried out by \textit{SIAC} and \textit{SEGUP};
    
    \item \textit{Statistics and Public Security Management Servers}: Professionals in the area of statistics and data science, knowledgeable about the criminology data used in the study, responsible for developing, together with multidisciplinary teams, strategies to combat violent crime;
\end{itemize}

It should be noted that among the 6 experts (two of each profile) appointed by the \textit{SIAC} board, there are experts with more than one profile (among those mentioned above), mainly because they are professionals with long careers in the area.

This research does not intend to identify each specialist profile in the results that will be presented later, so it will only refer to specialists as human 1 to human 6.

Each of the specialists was assigned the task of creating a feature relevance rank, but in order for this to become humanly feasible, instead of ranking the 35 input variables of the model, they were asked to rank 10 features only, since a total of 10 features is more easily to sort or even to classify conforms to the limits of an ordinal measurement scale for this task \cite{escala_ordinal_estatistica}. 

\begin{center}
\begin{figure}[h]
\includegraphics[scale=0.45]{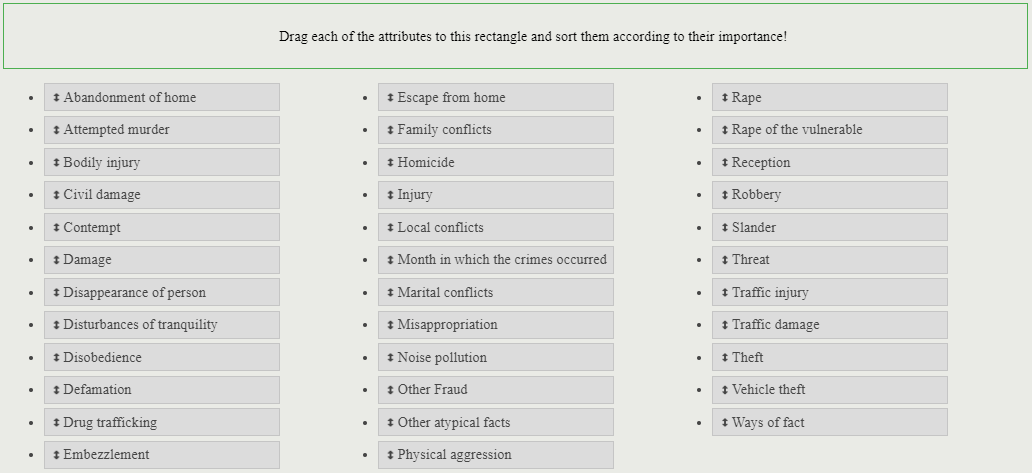}
\captionsource{Cutting from the web page uses information collection from experts, with a crime ranking that has not yet been created.}{\cite{ribeiro_pred2town_et_al_2021}.}
\label{fig_survey_clean}
\end{figure}
\end{center}

\begin{center}
\begin{figure}[h]
\includegraphics[scale=0.5]{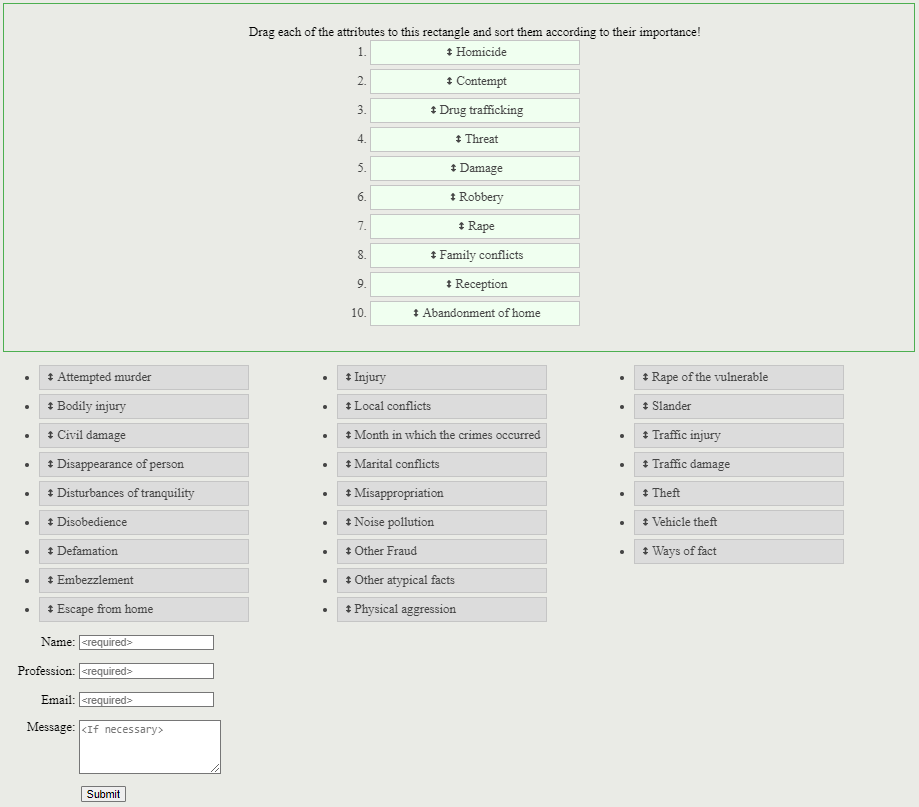}
\captionsource{Cutting from the web page uses information collection from experts, with a crime ranking already created.}{\cite{ribeiro_pred2town_et_al_2021}.}
\label{fig_survey_full}
\end{figure}
\end{center}

The task was performed using a computer, which presented a page to the specialist (\textit{``Availability of data and material''} item B.1). On this page, basic information about the context of the model was presented along with the research carried out, then the individual was asked to use the mouse and select 10 rectangles containing the names of the crimes that he believed to be most related to homicide crimes and place them in a queue, Figure \ref{fig_survey_clean} and \ref{fig_survey_full}.

During the rank process, in parallel, each expert assigned an ordinal value from 1 to 10 to each of the 10 features (in this case, the closer to 1 the feature is the more important it is). This enabled the creation of explanation ranks with 10 most important sub-deemed features for the homicide prediction process.

As described above, having in hand the ranks of XAI methods as well as the ranks generated by human experts, this research focused on working on ranks Correlation, Relation and Combinations analyzes to understand how Explanations and Expected of Interpretation are arranged.

\subsection{Rank Correlation and Relations}

As a quantitative way of comparing the ranks generated, whether by explanations of the XAI methods, or even by the ranks generated by human experts, we chose to use the \textit{Spearman Coefficient (SC)} \cite{spearman_ref}, with the aim of evaluating the similarities between the ranks, providing the values of the correlations and also assessments of the statistical significance (\textit{p-value}) of these correlations and values of \textit{Confidence Intervals} (\textit{ci}), as already known in the literature \cite{artusi2002bravais_confidence,bishara2017confidence_conf_notnormal,ferreira2015does_pvalue}, Table \ref{tab_spearman_range}.

\begin{table}[h!]
\begin{center}
\begin{tabular}{c|c}
\hline
\textbf{Range} & \textbf{Strength} \\ \hline
0.00 to 0.20       & Negligible        \\ \hline
0.21 to 0.40       & Weak              \\ \hline
0.41 to 0.60       & Moderate          \\ \hline
0.61 to 0.80       & Strong            \\ \hline
0.81 to 1.00       & Very strong       \\ \hline
\end{tabular}
\label{tab_spearman_range}
\captionsource{Intervals of the \textit{Spearman Correlation} metric and their Strength.}{\cite{spearman_ref}.}
\end{center}
\end{table}

However, before applying the \textit{SC} calculations, it was checked whether all the elements of a given rank $A$ existed in the rank $B$ to be compared, otherwise these elements were eliminated, and vice versa.

These eliminations were necessary, as the ranks generated by XAI methods have 35 positions, as the methods are capable of evaluating the relevance of feature for all of the model's input variables. However, the ranks generated by human experts only have 10 positions, due to human limitations in carrying out this task.

Thus, when calculating the correlations between two ranks corresponding to:

\begin{itemize}
    \item \textit{Two XAI Methods}: no elimination is made, as both ranks have a size equal to 35;
    \item \textit{One Expert and one XAI Method:} size 10 of the rank generated by the expert prevails and the elements present in the rank generated by the XAI method without correspondence are eliminated, preserving the positions of the other elements;
    
    \item \textit{Two Experts:} eliminations of elements are made in both ranks, since a given expert may not have selected a feature that another selected, therefore the size of the ranks may vary, with sizes such as 7, 8, 9 and 10 being found.
\end{itemize}

On the other hand, the relationships between the different ranks were also calculated quantitatively, but in a simpler way, through \textit{Set Theory}.

In this way, the ranks generated by XAI methods and the ranks generated by human experts were considered as two distinct sets, considering the first 10 attributes indicated by each of the XAI methods or even the experts.

This cut, considering only the first 10 features in each rank, is necessary for a fair set comparison. Since analyzes involving sets of elements do not consider the position in which they appear in a specific rank.

\subsection{Expectation of Interpretation}

The term Expectation of Interpretation or Expectation of Human Interpretation is proposed hereby in a generic way to refer to what is expected by the human expert as interpretations of the model explanations, this term is inspired by research in the area of XAI that questions the importance of human interaction with the explanations generated by XAI methods applied to black box models \cite{peterflach_humanintheloop,peter2020explainability,human_in_the_loop_zanzoto} and how the explanations of model are related to the expert's expectation of a problem by confronting their previous experiences and also the confidence in the model being analyzed \cite{kaur2020interpreting,hall2017machine,hong2020human,expectation_ronny}.

Thus, to generate the expectation of interpretation, the expert analyzes the machine learning model input data (features) and its performance (for example, accuracy), considers their pre-knowledge of the proposed problem (context) by re-signifying the use of each feature, and finally performs the creation of what is expected as a possible explanation to the model (for example, rank of relevance of features) based on their interpretation of what is considered more consistent for the problem.

In other words, the expectation of interpretation of the machine learning model is an explanation created by a human expert who is based upon their interpretation of the context.

It is important to point out that, in order to carry out the process of creating the expectation of interpretation advocated above, this research deems as important that the expert has no previous contact with the results of the explanations generated by the XAI methods, so as to avoid the creation of biases.

Another important highlight is that the expectation of interpretation does not emerge as a universal explanation proposal for a model, but can be easily complemented by explanations generated through XAI methods, given the capability of these tools. 

In this sense, this research seeks to introduce a proposal to combine global feature relevance rankings generated from XAI methods to the rankings obtained from the application of the expectation of interpretation, tool called \textit{ConeXi}.

This technique aims at providing the inclusion of the human being in the loop of the process of opening the black box models, since in a next step it can enable the combination of results obtained by XAI methods and human experts. 

\subsection{Combining Explanation and Expectation of Interpretation}

\textit{Spearman Correlation} and \textit{Set Theory} fundamentals let us compare different ranks derived from XAI measures and interpretability expectations. 

Another interesting and innovative way to use the results generated both by XAI methods and by human experts is to combine the explanations results with the interpretations results, with the purpose of guaranteeing the human participation in the process of opening the black box models, called by this research \textit{Combining Explanations and Interpretations (ConeXi)}.

The processes carried out by \textit{ConeXi} are defined by the equations \ref{eqn:nota_I} and \ref{eqn:nota_II}, which together represent the basic mathematical process performed to calculate scores for each feature and thus combine the ranks of explanation and interpretation, ultimately aiming at creating a single overall ordered rank where features closer to 1 have higher explanation and interpretation.

Equations \ref{eqn:nota_I} and \ref{eqn:nota_II} represent a $S$ computation system applied to each feature in the ranks. For this, you need to consider the equations and their variables:

\begin{equation}
\label{eqn:nota_I}
S = c + \sum_{i=m}^{n} f_{i}
\end{equation}

where $c$ is the computation required to create an intermediate table containing the numerical positions of each feature in the ranks. Then we have $n$ as the number of features in each rank (equal to 35) and $m$ as the initial index of the iteration (equal to 1). In simplistic terms, $f_i$ is the number of times a given feature $i$ appears in a specific ranking position (with some observations in the count).

For better understanding, $f_i$ can be represented as a equation \ref{eqn:nota_II}:

\begin{equation} 
\label{eqn:nota_II}
f_i = \sum_{j=q}^{r}\sum_{k=s}^{t} p_{jk} * w_{i}
\end{equation}

where $r$ represents the number of ranks existing in the analyzes (equal to 12, i.e., 6 by the XAI methods and 6 by experts) and $q$ is the initial index of the iteration (equal to 1). Which carry out a second process from $k$ (equal to 1) to $t$ (equal to 10), with $t$ being a user-defined constant. In $p_{jk}$ it is increased by 1 if the analyzed attribute $i$ is in a position lower than or equal to the limit position of the combination $k$ in rank $j$, otherwise no increment occurs.

As above, in each rank, only feature positions less than or equal to $k$ will be considered during the iteration process. In other words, $t$ is an important parameter to be passed to \textit{ConeXi}, as it serves as a limit for the features of each rank in the combination process.

Lastly, the value of $w_i$ is the weight that will multiply the score of a specific feature $i$, thus maximizing or minimizing the relevance of a specific feature, since the weight value can be manipulated by a human user seeking to explain the model (herein using $w_j$ = 1, for all features).

After calculating the $S$ value for all the features in the analyzed ranks, one can simply sort the $S$ values in descending order and thus find an overall rank that is the combination of the explanation ranks and the interpretation ranks.

Aiming to facilitate the understanding of how \textit{ConeXi} works, below is the execution of the processes presented in the equations \ref{eqn:nota_I} and \ref{eqn:nota_II}, in a simple example of combining the 7 features (A, B, C, D, E, F, and G) present in 3 different ranks, considering only the first 4 elements of each rank table \ref{tab_conexi_1}.

\begin{table}[h!]
\begin{center}    
\begin{tabular}{c|c|c}
\hline
Rank 1 & Rank 2 & Rank 3 \\ \hline
A      & A      & A      \\ \hline
B      & B      & C      \\ \hline
C      & D      & B      \\ \hline
D      & C      & F      \\ \hline
E      & G      & D      \\ \hline
F      & E      & E      \\ \hline
G      & F      & G      \\ \hline
\end{tabular}
\end{center}
\caption{Example of three ranks containing elements A, B, C, D, E, F and G in different positions.}
\label{tab_conexi_1}
\end{table}

At first, \textit{ConeXi} creates an intermediate table, containing the position of each of the features $i$ to be iterated in the different ranks $j$, table \ref{tab_conexi_2}.

\begin{table}[h!]
\begin{center}
\begin{tabular}{c|c|c|c}
\hline
Iterations & j = Rank 1 & j = Rank 2 & j = Rank 3 \\ \hline
i = A      & 1          & 1          & 1          \\ \hline
i = B      & 2          & 2          & 3          \\ \hline
i = C      & 3          & 4          & 2          \\ \hline
i = D      & 4          & 3          & -          \\ \hline
i = E      & -          & -          & -          \\ \hline
i = F      & -          & -          & 4          \\ \hline
i = G      & -          & -          & -          \\ \hline
\end{tabular}
\caption{Intermediate table containing the positions of each feature to be iterated in the ranks.}
\label{tab_conexi_2}
\end{center}
\end{table}

Then, based on the table \ref{tab_conexi_2}, \textit{ConeXi} counts how many times a certain feature from iteration $i$ appears in a position less or equal to a specific position $k$ in the rank, generating a new intermediate table that is the extension of the previous one, table \ref{tab_conexi_3}.

Finally, the value of $S$ is found from the sum of the scores calculated in table \ref{tab_conexi_3} column $S$. Remembering that at this moment, different weights (in percentage format) multiply the scores obtained by each feature according to the interest of a human individual (ensuring human participation in the results). 

\begin{table}[h!]
\begin{center}
\begin{tabular}{|c|c|c|c|c|c|c|c|c|}
\hline
Iterations & j = Rank 1 & j = Rank 2 & j = Rank 3 & k\textless{}=1 & k\textless{}=2 & k\textless{}=3 & k\textless{}=4 & S    \\ \hline
i = A      & 1          & 1          & 1          & 3              & 3              & 3              & 3              & $12*w_i$ \\ \hline
i = B      & 2          & 2          & 3          & 0              & 2              & 3              & 3              & $8*w_i$  \\ \hline
i = C      & 3          & 4          & 2          & 0              & 1              & 2              & 3              & $6*w_i$  \\ \hline
i = D      & 4          & 3          & -          & 0              & 0              & 1              & 2              & $3*w_i$  \\ \hline
i = E      & -          & -          & -          & 0              & 0              & 0              & 0              & $0*w_i$  \\ \hline
i = F      & -          & -          & 4          & 0              & 0              & 0              & 1              & $1*w_i$  \\ \hline
i = G      & -          & -          & -          & 0              & 0              & 0              & 0              & $0*w_i$  \\ \hline
\end{tabular}
\caption{Count and sum $S$ of feature positions in ranks.}
\label{tab_conexi_3}
\end{center}
\end{table}

In this example, only the first 4 features of each rank were considered, and with the combination between ranks the output of \textit{ConeXi} may present more than 4 features, given the difference in features existing in each rank.

Thus, the final value of $S$ is the score that defines the position of a given feature in the final ranking (the higher the score, the closer the feature is to position 1). In other words, for the example in question, the combination generates the rank A, B, C, D, and F.

Note that the final rank generated in the \ref{tab_conexi_3} table makes full sense, since for the feature:
\begin{itemize}
    \item[A:] presents a score equal to 12, because in the iterations of $k$ it received scores 3, 3, 3, and 3, as it appeared in first position in all ranks;
    
    \item[B:] presents a score equal to 8, because in the iterations of $k$ it received scores 0, 2, 3, and 3, as it appeared in second position in two ranks and in third position in one rank;
    
    \item[C:] presents a score equal to 6, because in the iterations of $k$ it received scores 0, 1, 2, and 3, as it appeared in the second, third and fourth position of each rank;

    \item[D:] presents a score equal to 3, because in the iterations of $k$ it received scores 0, 0, 1 and 2, as it appeared in the fourth and third position of two ranks;

    \item[E:] presents a score equal to 0, as it does not appear in a position less than or equal to 4 in any of the ranks;
    
    \item[F:] presents a score equal to 1, because in the iterations of $k$ it received scores 0, 0, 0, and 1, as it appeared in fourth position in only one rank;

    \item G: presents a score equal to 0, as it does not appear in a position less than or equal to 4 in any of the ranks.
    
\end{itemize}

\textit{ConeXi} was born from the need to combine different ranks of elements into a single final rank, based on the frequency that certain elements appear in specific positions in the ranks. Therefore, this tool allows three different forms to combinations to create the final ranking:

\begin{itemize}
    \item \textit{No human interaction:} in this form, simply the ranks generated by different XAI methods are combined with each other, as in an ensemble. There is no human interaction through weights or even through rankings of expectations;
    
    \item \textit{With human interaction by weights:} in this form, the human individual declares different weights to the elements of the ranks, thus modifying the relevance of each element (also understood as a feature);
    
    \item \textit{With human interaction through the addition of ranks:} in this way, the human individual adds ranks created by experts, together with the ranks generated by XAI methods.
\end{itemize}

This research specifically uses the last form of the rank combination process. Since one of the purposes here is to allow the human in the loop of the process of explaining machine learning models.

More about \textit{ConeXi} can be see in section \textit{``Code availability''} item A.3.

\section{Results and Discussion}

This section will present the results of the research divided into 4 different parts: The section \ref{preparation}, the strategy used to select the ideal test split that will be the basis for the explanation process will be presented; The section \ref{correlations}, a comparative analyzes carried out through correlation calculations of the generated ranking pairs will be presented; In the section \ref{relations}, a comparative analyzes involving different sets of ranks will be presented; Finally, on section \ref{combinations}, the result of the combination of all ranks will be presented.

In the sections \ref{correlations}, \ref{relations}, and \ref{combinations}, results will be presented that seek to answer, through analyzes based on clues and evidence, the questions set out in Figure \ref{fig_perguntas}:

\begin{figure}[h]
\includegraphics[width=\textwidth]{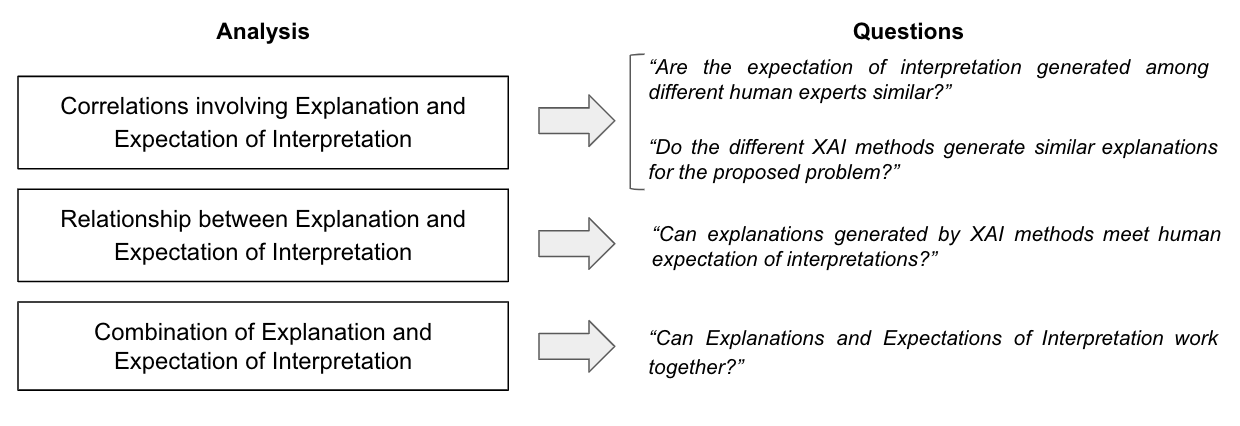}
\caption{Mapping of Analyzes vs. Questions to be answered.}
\label{fig_perguntas}
\end{figure}

\subsection{Selection of the test split to be explained}\label{preparation}

Once the \textit{RF} algorithm was properly tuned, trained, tested and evaluated, in accordance with what was advocated in \cite{ribeiro_pred2town_et_al_2021}, we sought to inspect the \textit{Cross Validation with Stratification} process based on a quantity of $folds = 7$, seeking to evaluate the accuracy of the models created from the fold combinations of each split with the objective of selecting the ideal split of test to be explained by XAI methods.

In Table \ref{tab_folds}, the columns ``Folds'' represents the number of folds into which the dataset was divided, while the lines ``Split'' represents different models tests with a specific fold $i$. Note, if a model was tested with a fold $i$, for example, it means that it was trained with the other folds existing in the same column.

In this experiment, the number of folds in the dataset was varied, with values from 3 to 10 folds. It was verified that the average accuracy varied little between the tests (staying between 0.73 and 0.75). In this case, we chose to make focus in 7 Folds, since the model had already been selected (among other algorithms) using this same number of folds in \cite{ribeiro_pred2town_et_al_2021}, Table \ref{tab_folds} column ``Folds'' 7.

\begin{table}[]
\centering
\resizebox{\textwidth}{!}{
\begin{tabular}{c|c|c|c|c|c|c|c|c}
\hline
\textbf{Model test} & \textbf{Folds = 3} & \textbf{Folds = 4} & \textbf{Folds = 5} & \textbf{Folds = 6} & \textbf{Folds = 7} & \textbf{Folds = 8} & \textbf{Folds = 9} & \textbf{Folds = 10} \\ \hline
\textbf{Split 1} & 0.73 & 0.72 & 0.70 & 0.70 & 0.72 & 0.70 & 0.74 & 0.71 \\ \hline
\textbf{Split 2} & 0.75 & 0.76 & 0.75 & 0.77 & 0.73 & 0.75 & 0.71 & 0.70 \\ \hline
\textbf{Split 3} & 0.73 & 0.75 & 0.78 & 0.79 & 0.74 & 0.73 & 0.77 & 0.75 \\ \hline
\textbf{Split 4} & - & 0.70 & 0.74 & 0.73 & 0.81 & 0.80 & 0.76 & 0.76 \\ \hline
\textbf{Split 5} & - & - & 0.71 & 0.74 & 0.75 & 0.75 & 0.80 & 0.80 \\ \hline
\textbf{Split 6} & - & - & - & 0.68 & 0.73 & 0.75 & 0.72 & 0.73 \\ \hline
\textbf{Split 7} & - & - & - & - & 0.70 & 0.71 & 0.75 & 0.76 \\ \hline
\textbf{Split 8} & - & - & - & - & - & 0.70 & 0.73 & 0.73 \\ \hline
\textbf{Split 9} & - & - & - & - & - & - & 0.68 & 0.70 \\ \hline
\textbf{Split 10} & - & - & - & - & - & - & - & 0.69 \\ \hline
\textbf{Mean} & 0.74 & 0.73 & 0.74 & 0.74 & \textbf{0.74} & 0.74 & 0.74 & 0.73 \\ \hline
\end{tabular}}
\caption{Accuracy of \textit{RF} executions with different amounts of folds and splits.}
\label{tab_folds}
\end{table}

This analysis made it possible to identify the accuracy values of the models created from the process of training and testing the combinations of the 7 folds. In other words, if the accuracy value found from a test carried out with a fold $i$ is low (for example, 0.70 in line ``Split'' = 7 and column ``Folds'' = 7), it means that that test set is more difficult for the model predict than the others, the opposite idea is being also true (for example, when considering the accuracy 0.81 of the line ``Split'' = 4 and column `Folds'' = 7).

Aiming to produce more solid explanations regarding the analyzed model and minimize the problem described above, this research chose to work with the model coming from tests with ``Split 3''  created by division of dataset into 7 folds, Table \ref{tab_folds} (line ``Split'' = 3 and column ``Folds'' = 7), given that it presents its accuracy value close to the average 74 of the models analyzed in the experiment, Table (line ``Mean'' and column ``Fods'' 7).

In detail, this research used the strategy presented in this section as a way of better understanding the model according to its performance, highlighting that even with very close accuracy values between the different tests, it was possible to identify an ideal test to carry out the explanations.

Therefore, in subsequent sections the results that will be presented are related to this model that was trained with folds = 1, 2, 4, 5, 6, and 7 and tested with fold 3.

All analyzes carried out using the model originating from the chosen split were also replicated for the other models generated in the other splits of the dataset division into 7 folds, and can be viewed in the section \textit{``Code availability''} item A.1.

\subsection{Correlations involving Explanations and Expectation of Interpretation} \label{correlations}

\begin{figure}[h]
\begin{center}
\includegraphics[scale=0.7]{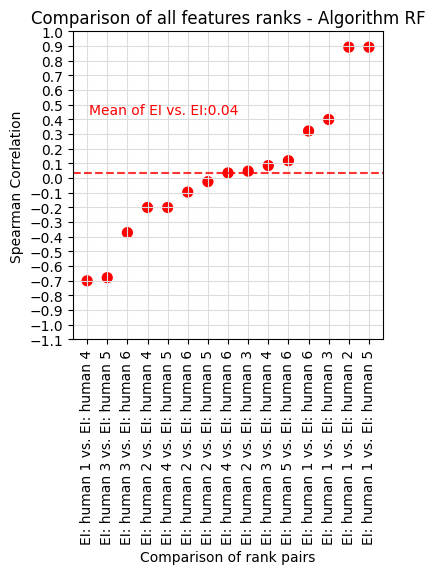}
\caption{Spearman Correlation calculated for each rank pair generated by Expectation of Interpretations (EI).}
\label{fig_ei_vs_ei}
\end{center}
\end{figure}

\begin{figure}[h]
\begin{center}
\includegraphics[scale=0.7]{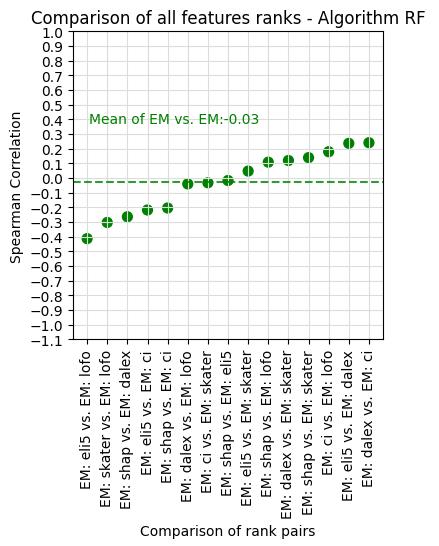}
\caption{Spearman Correlation calculated for each rank pair generated by Explanation Methods (EM).}
\label{fig_em_vs_em}
\end{center}
\end{figure}

\begin{figure}[h]
\begin{center}
\includegraphics[scale=0.5]{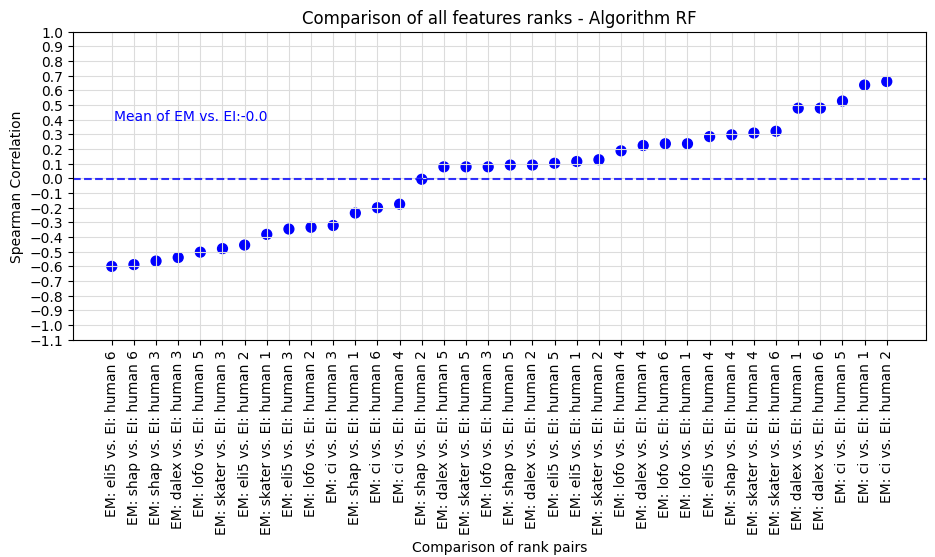}
\caption{Spearman Correlation calculated for each rank pair generated by Explanation Methods (EM) and Expactation of Interpretations (EI).}
\label{fig_em_vs_ei}
\end{center}
\end{figure}

\begin{figure}[h]
\begin{center}
\includegraphics[scale=0.65]{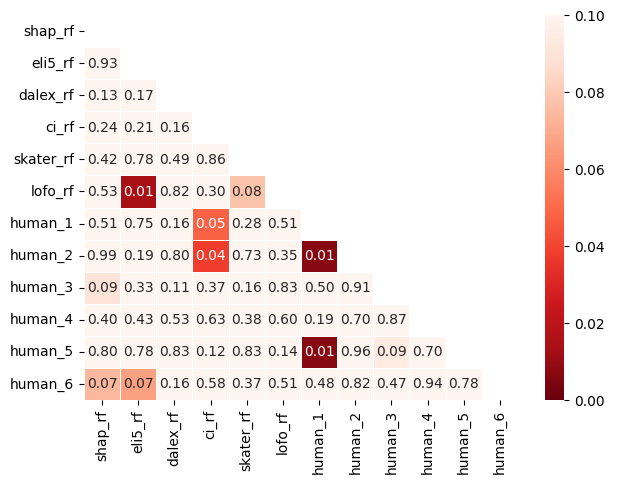}
\caption{Statistical significance (\textit{p-value}) found through \textit{Spearman} correlation, for each rank pair.}
\label{fig_spearman_p}
\end{center}
\end{figure}

The discussions arising from the analyzes of Figures \ref{fig_ei_vs_ei}, \ref{fig_em_vs_em}, and \ref{fig_em_vs_ei}, used information on the statistical significance values present in Figure \ref{fig_spearman_p}, aiming to provide a statistical basis for the statements performed using the \textit{p-value} (obtained from \textit{Spearman Correlation}), and \textit{Confidence Interval ci} (obtained from $ci = (1 - $\textit{p-value}) * 100).

As a first analysis involving ranks, we present the \textit{Spearman} correlations calculated between all pairs of Expectation of Interpretation - EI ranks generated by human experts, Figure \ref{fig_ei_vs_ei}.

In Figure \ref{fig_ei_vs_ei}, the $x$ axis shows the pairs of EI ranks, while the $y$ axis shows the values of the \textit{Spearman} correlations calculated for each pair of ranks (red dots). It is still possible to observe the general average correlation (red dashed line) at 0.04, considered ``weak''.

In this first comparative analysis, Figure \ref{fig_ei_vs_ei}, the highest correlations are observed in ``EI: human 1 vs. EI: human 5'' and ``EI: human 1 vs. EI: human 2'', with respectively 0.88 and 0.99 correlations (both with: \textit{p-value} $= 0.01$ $\therefore$ $ci = 99\%$ of confidence), considered ``very strong''. Meaning an alignment of existing interpretation expectations between the 3 different specialists.

Also, observing the same figure, the comparisons ``EI: human 1 vs. EI: human 4'' and ``EI: human 3 vs. EI: human 5'' presented the highest negative correlations respectively -0.70 (with \textit{p-value} $= 0.19$ $\therefore$ $conf= 81\%$) and 0.68 (with \textit{p-value} $= 0.09$ $\therefore$ $ci = 91\%$), considered ``Strong''. Meaning that the compared ranks present an inverse correlation, that is, the experts involved in the comparisons gave inverse relevance to the features existing in the compared ranks.

The other comparisons obtained correlations between 0.4 and -0.4, considered ``weak''. Meaning a low agreement between related ranks, Figure \ref{fig_ei_vs_ei}.

This research attributes the considerable number of ``weak'' comparisons, Figure \ref{fig_ei_vs_ei}, to the different professional profiles, life experiences and individual perspectives of the experts involved in the study.

These results show the difficulties that exist in the model interpretation process, because as seen above, not even the experts' interpretations completely agree with each other, so wanting the explanations of XAI methods to meet the experts' expectations proves to be a task and high complexity.

The results above show that even for a human expert, the action of explaining a learning model and the context in which it is inserted is not an easy task, because if the opposite were true, we would have greater correlations between the pairs of ranks analyzed. This is the result for the previously asked question: ``\textit{Are the expectation of interpretation generated among different human experts similar?}''.

When observing the relevance ranks generated by different Model Explanation Methods (EM), Figure \ref{fig_em_vs_em}, it can be noted that none of the correlations have a value above 0.4, verifying that in the vast majority of comparisons the result was a positive or even negative ``weak'' correlation (values between 0.4 and -0.4).

Still observing the correlations present in Figure \ref{fig_em_vs_em}, it is also possible to highlight the negative correlations found for the comparisons ``EM: eli5 vs. EM: Lofo'' and ``EM: skater vs. EM: lofo'', which respectively presented correlations -0.4 (with \textit{p-value} $= 0.01$ $\therefore$ $ci = 99\%$) and -0.3 (with \textit{p-value} $= 0.08 $ $\therefore$ $ci = 92\%$), considered ``weak''.

These results, obtained from Figure \ref{fig_em_vs_em}, show that the machine learning model being explained is difficult to explain, as defended in \cite{ribeiro_complexity_et_al_2021}. In other words, for a difficult model, it is expected that the XAI methods used in the explanation process disagree with each other, given the complexity of the problem that the model generalizes.

\begin{figure}[!h]
\begin{center}
\includegraphics[scale=0.77]{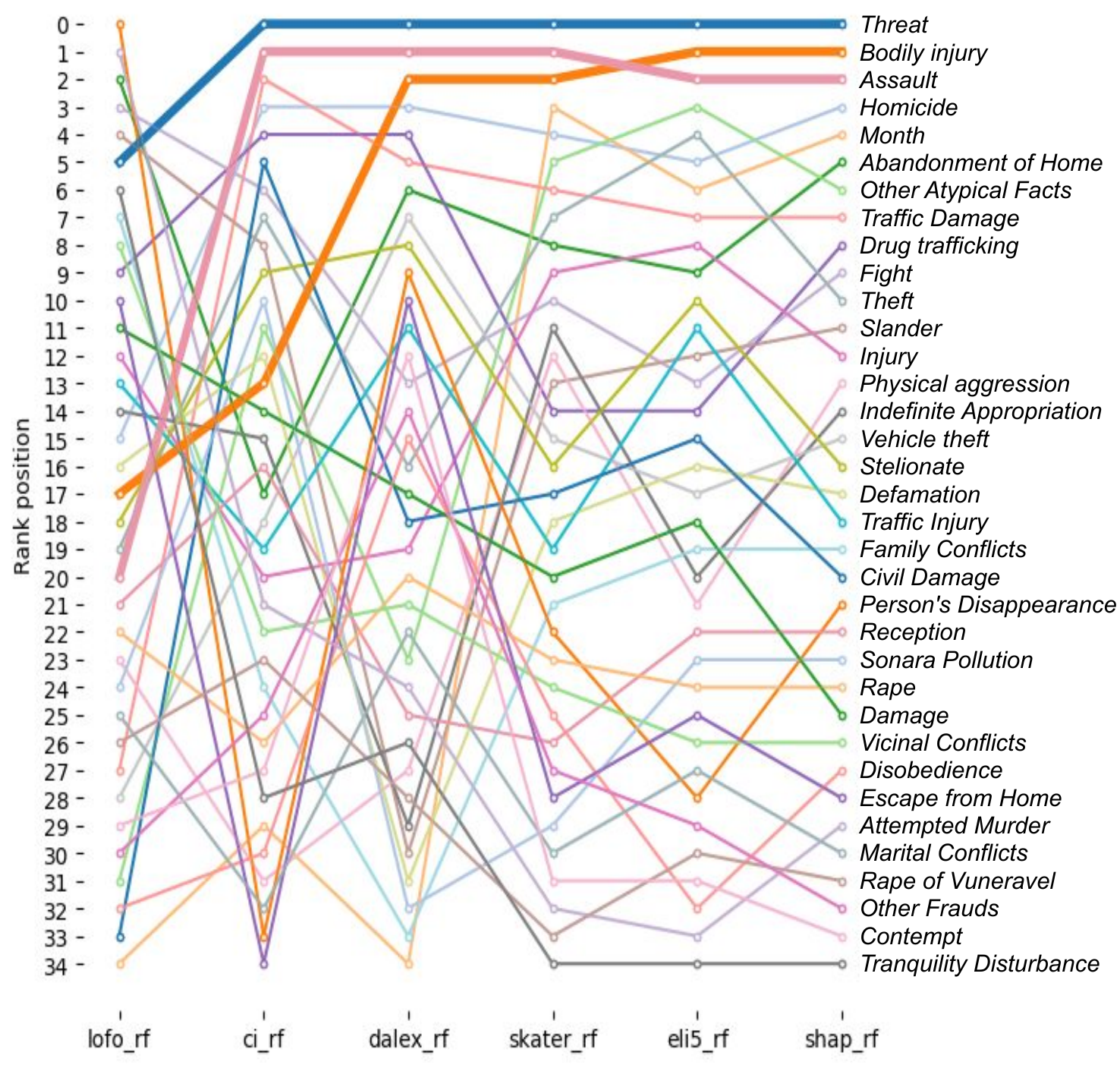}
\caption{Position of each feature in the explanation ranks generated by the XAI methods.}
\label{fig_bump_chat}
\end{center}
\end{figure}

An interesting way to visualize the ranks generated by XAI methods is the visual comparison of each feature relevance rank, which even for ranks with low correlations, it is possible to observe relevance trends of specific features, Figure \ref{fig_bump_chat} .

In Figure \ref{fig_bump_chat}, on the $y$ axis, the positions of each feature present in the attribute relevance ranks of the different ranks are presented. On the $x$ axis, the 6 ranks of explanations of the models generated by the XAI methods are presented. Note, the further to the right an XAI method appears, the greater its correlation with the neighboring rank (based on the correlation values presented in Figure \ref{fig_em_vs_em}).

Thus, given the low correlations found between the relevance ranks of attributes visualized in Figure \ref{fig_bump_chat} (thicker lines), we can point out the features ``Threat'', ``Bodily injury'' and ``Assault '', as the 3 features most indicated by most XAI methods (\textit{Shap, Eli5, Skater,} and \textit{Dalex}) that best explain the analyzed model. These analyzes are the answer to the question: ``\textit{Do the different XAI methods generate similar explanations for the proposed problem?}''.

When comparing the EM and EI ranks, Figure \ref{fig_em_vs_ei}, it is possible to point out 5 comparisons that obtained a correlation greater than 0.4, they were ``Em: ci vs. EI human 2'' (\textit{p-value} $= 0.04$ $\therefore$ $ci = 96\%$), ``EM: ci vs. EI: human 1'' (\textit{p-value} $= 0.05$ $\therefore$ $ci = 95\%$), ``EM: ci: vs. EI: human 5'' (\textit{p-value} $= 0.12$ $\therefore$ $ci = 88\%$), ``EM: dalex vs. EI: human 6'' (\textit{p-value} $= 0.16$ $\therefore$ $ci = 84\%$), and ``EM: dalex vs. EI: human 1'' (\textit{p-value} $= 0.16$ $\therefore$ $ci = 84\%$), that is, it can be said that the XAI methods called \textit{CI} and \textit{Dalex} were the ones that best met the interpretation expectations of at least 4 experts with ``moderate'' strength.

With the exception of the 5 comparisons mentioned in the previous paragraph, the other comparisons shown in Figure \ref{fig_em_vs_ei} present positive and negative ``weak'' values, or even ``moderate'' negative values.

The results above shed light on an answer to the question ``\textit{Can explanations generated by XAI methods meet human expectation of interpretation?}'', but they are not yet definitive and the answer to this question will be further explored in the next section.

Seeking to present a summary of the statistical significance of the \textit{Spearman} correlations calculated for all rank comparisons, Figure \ref{fig_spearman_p} shows all \textit{p-values} found in this experiment.

In Figure \ref{fig_spearman_p}, the rows and columns of the diagonal matrix show the different pairs of ranks compared. The matrix cells are the \textit{p-values} found in each comparison.

When evaluating only statistically significant values, Figure \ref{fig_spearman_p} cells with \textit{p-values}$<=0.05$ ($ \therefore ci = 95\%$), attention is drawn to the comparisons:
\begin{itemize}
    \item ``human 1'' vs. ``human 5'' and ``human 1'' vs. ``human 2'': which present positive correlations with ``very strong'' strength (\ref{fig_ei_vs_ei}), indicating the statistical significance of the alignments of the 3 experts involved;
    \item ``ci'' vs. ``Human 1'' and ``ci'' vs. ``Human 2'': which presents the \textit{CI} method as being the method that met the expectations of interpretations with ``strong'' strength and statistical significance for two human experts. This XAI method is the most suitable for explaining the model within the context of predicting homicide crimes;
    \item ``eli5'' vs. ``lofo'': which presented a -0.4 correlation and statistical significance, indicating that its ranks present inverse correlations with ``moderate'' strength.
\end{itemize}

\subsection{Relationship between Explanations and Expectation of Interpretations} \label{relations}

As a relationship between Explanations and Expectation of Interpretations, comparative analyzes are defined as performed by creating two main sets, one Set of Explanation Methods (\textit{SEM}) and another Set of Expectation of Interpretations (\textit{SEI}), substantiating the results on the context of the proposed problem. This, considering only the first 10 features of all generated ranks.

As previously mentioned, it is important to emphasize that at no time does this research intend to question the features pointed out by specialists in the area of security, because each of these features was consciously indicated by professionals who deal directly or indirectly with data related to the confrontation of violent crimes in the city of Belém, in the state of Pará.

\begin{figure}[h]
\begin{center}
\includegraphics[scale=0.35]{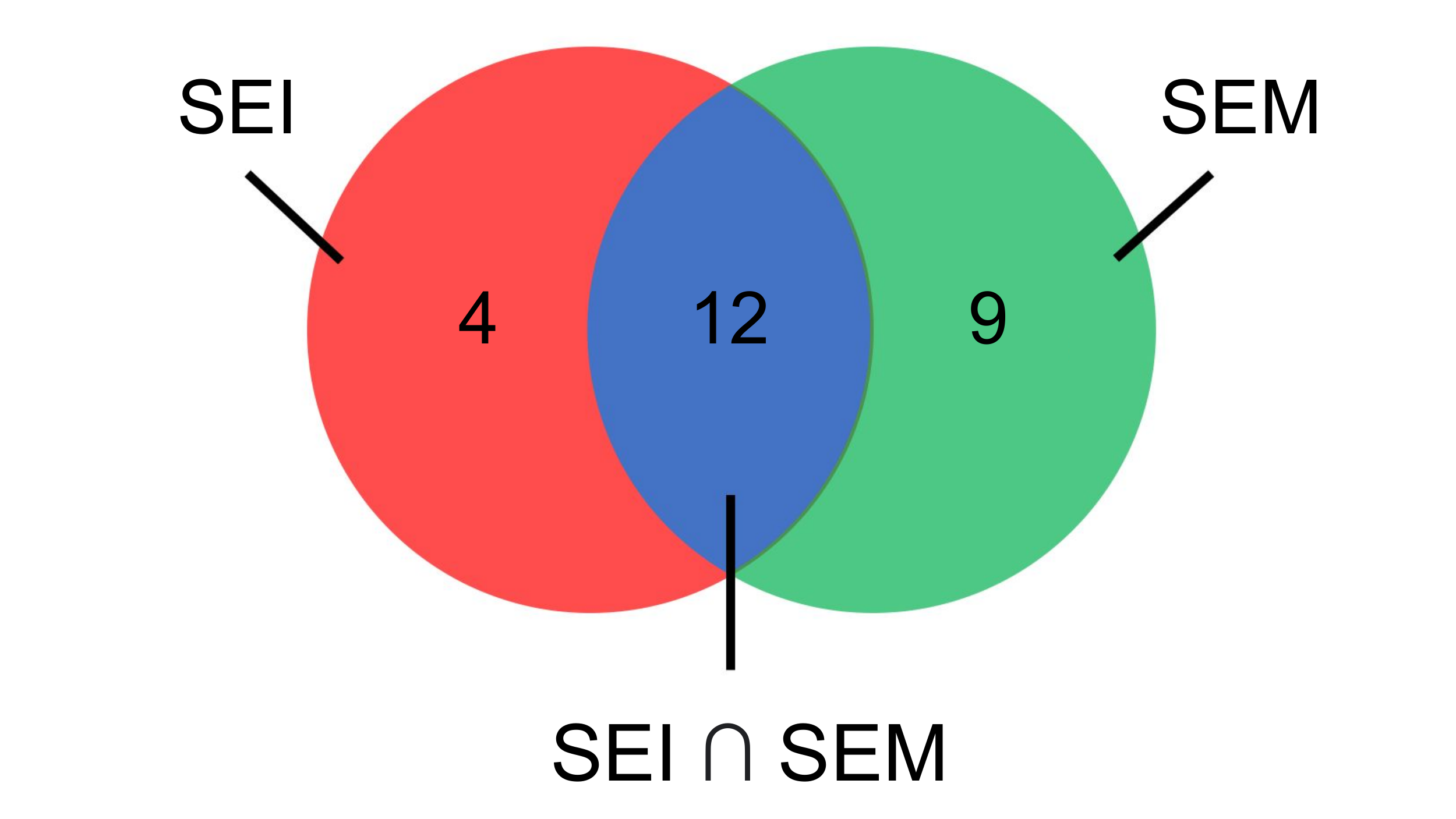}
\end{center}
\caption{Graphical representation of the two sets of features created: Set of Expectation of Interpretations (SEI) and Set of Explanation Methods (SEM)}
\label{fig_conjuntos}
\end{figure}

In this regard, two large sets, \textit{SEM} and \textit{SEI}, were created without the presence of feature repetitions, and the intersection between these two large sets were calculated, the result being shown in Figure \ref{fig_conjuntos}.

Figure \ref{fig_conjuntos} clearly shows that the \textit{SEI} set has a total of 16 features and the \textit{SEM} set has 21 features. These two sets have an intersection of 12 features, so they agree with an approximate percentage of 49\% of the total of features.

The results above show that despite the high number of features indicated by the explanations methods (21 features in total), there are features from the set of expectation of interpretations that were not related. So, this shows that approximately 75\% of the expectation of interpretations of the experts were met (considering 16 \textit{SEI} features = 100\%), thus answering the question posed earlier in this paper: \textit{``Can explanations generated by XAI methods meet human expectation of interpretations?''}.

The considerable number of features that exist exclusively in SEM (9 in total) demonstrates that the XAI methods point to features that are not being considered as priorities to human experts, thus, indicating that experts can explore even more explanatory variables of the model.

Table \ref{tab_agrupamentos} deals with the context of each feature present in the sets created.

\begin{table}[!h]
\centering
\caption{Atributes in the SEI and SEM}
\resizebox{\textwidth}{!}{%
\begin{tabular}{|c|c|c|c|}
\hline
Set & $SEI - SEM$ & $SEI \cap SEM$ & $SEM - SEI$\\ \hline
Quantity of features & 4 & 12 & 9 \\ \hline
Name of the attibutes & \multicolumn{1}{l|}{\begin{tabular}[c]{@{}l@{}}
- Vulnerable Rape;\\ 
- Rape;\\ 
- Physical Aggression;\\ 
- Marital Conflict.
\end{tabular}} & \multicolumn{1}{l|}{\begin{tabular}[c]{@{}l@{}}

- Vicinal Conflicts;\\
- Assault;\\
- Homicide;\\
- Threat;\\ 
- Attempted Murder;\\
- Fight;\\
- Drug Trafficking;\\
- Month;\\
- Person Disappearance;\\
- Vehicle Theft;\\
- Bodily Injury;\\
- Family Conflicts.\\

\end{tabular}} & \multicolumn{1}{l|}{\begin{tabular}[c]{@{}l@{}}
- Stelionate;\\
- Tranquility Disturbance;\\ 
- Injury;\\
- Traffic Damage;\\ 
- Slander;\\ 
- Other Atypical Facts;\\ 
- Theft;\\ 
- Civil Damage;\\ 
- Home Abandonment.
\end{tabular}} \\ \hline
\end{tabular}%
}
\label{tab_agrupamentos}
\end{table}

This research understands that the results presented in the set of expectation of interpretation, $(SEI - SEM) + (SEI \cap SEM)$ refer to Table \ref{tab_agrupamentos}, are features strongly related to the proposed problem and to human interpretation, since they were directly indicated by experts who deal with crime-related data in their daily lives and have relevant knowledge in the public security area due to their professional backgrounds. Therefore, it is not for this research to question the applicability in the context or even the relevance of such features, considering these features as the ideal of model explanations through the human perspective.

An important observation about the $SEI - SEM$ results shown in Table \ref{tab_agrupamentos}, is that the 4 crimes exclusively indicated by human experts (``Vulnerable Rape'', ``Rape'', ``Physical Aggression'' , and ``Marital Conflict'') are sensitive crimes to discuss because they have a high emotional nature along with a broad social appeal.

The results presented in the column $SEM - SEI$ point to evidence of the existence of a high number of crimes (9 crimes in total) that, according to the proposed model and the explanations that were generated, can explain the process of homicide prediction and can be considered and analyzed by specialists in the field of criminology, so as to allow for a better and broader understanding of the dynamics of crime occurrences in the city under study and new interpretations of the model.

\subsection{Combination of Explanations and Expectation of Interpretations} \label{combinations}

\begin{figure}[h]
\begin{center}
\includegraphics[scale=0.46]{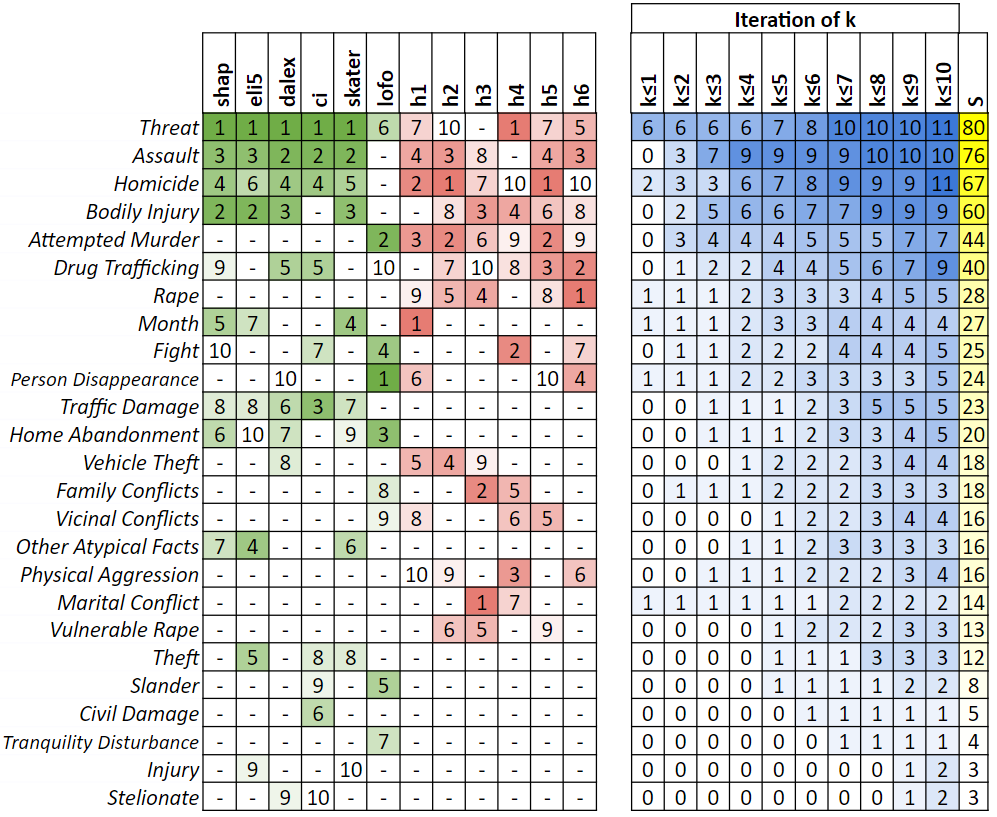}
\end{center}
\caption{Heatmap indicating the position (numbers 1 to 10) of each model input feature in a specific rank (columns).}
\label{fig_heatmap}
\end{figure}

As a Combination of Explanations and Expectation of Interpretations, a set of analyzes are defined that seek to facilitate the combination of the different ranks existing in this study. To this end, this research applied the proposed ranking combination presented earlier (called \textit{ConeXi}) to the collected results, and thus constructed an overall ranking visualization that demonstrates the combination of efforts of XAI methods with human experts, Figure \ref{fig_heatmap}.

The illustration shows the dynamics of the positions of each feature indicated by explanation methods (heatmap green) with the features indicated the expectation of interpretations (heatmap red) generated by experts. Still in Figure \ref{fig_heatmap} there is part referring to iterations (heatmap blue) calculate the features points in the proposed algorithm, and finally the $S$ score is calculated in the last heatmap (heatmap yellow).

Building up Figure \ref{fig_heatmap} required a sorting of the lines according to the values of $S$, according to \textit{ConeXi}'s internal procedures, shown earlier. Therefore, at each iteration of $k$, the number of times a given feature appears in the ranks in a position $k<=x$ was increased, where $x$ varies from 1 to 10, Figure \ref{fig_heatmap} Iteration table to the right. 

In the next step, the Sum of Points, the scores of each feature are added in $S$. Finally, they are sorted in descending order (Figure \ref{fig_heatmap} Iteration table, column $S$.

The results presented in the herein study are indicative that both current XAI methods can be improved to come even closer to how human experts explain and interpret the model and the context of the problem. But, also, the results show that experts can consider the explanations generated by XAI methods and enrich their understandings about the context/problem and the analyzed model concerned. Both results can be used in a single Collaborative Explanation, thus answering the third and last question posed hereby: ``\textit{Can Explanations and Expectation of Interpretations work together?}''.

Therefore, according to all the results presented herein, it is considered that a Collaborative Explanation between Man and Machine may be the best way to succeed in the opening process of black box models, as even with the great potential and analytical capacity of XAI methods, the expectation of interpretations represent results that complement the results pointed out by XAI methods, thus providing new inputs for discussions concerning the context to which the machine learning model is inserted. 

\section{Final Considerations}

However, given the sensitive context of the problem of predicting homicides in the real world, together with the various explanations and expectations of interpretations of this model, it was noted that human expectations were not fully met at any time, thus revealing that XAI methods appear not to be user-centric. This, at the same time that these methods must point out characteristics that explain the model and the context in which it is inserted through new perspectives.

The methodology of combining expectation positions of interpretations and explanations shows promise, as it allows combining machine results and human results and thus tends to overcome the differences identified between these two groups, but this technique is limited to interpretations and explanations based on rank structure.

The research results represent the difficulties in explaining the decisions of black box algorithms in real scenarios and highlight the need for more studies and methods in this area. Emphasizing the need to have a human expert inserted in the process of explaining machine learning models in order to build a more acceptable collaborative explanation between man and machine.

\section{Future works}

In view of the foregoing, in the next steps of this research, validating the concept of interpretability expectation for other contexts is intended in order to expand the understanding thereof to the XAI community. Also, proposing an explainability method that provides a new way to explain machine learning models is intended as based mainly on data complexity through new techniques grounded on Item Response Theory (IRT) to add efforts to the existing XAI methods project.

\section{Acknowledgements}

The researchers hereof would like to thank the Secretariat of Intelligence and Criminal Analyzes (SIAC), in the city of Belém, Pará State, which provided its specialists to voluntarily participate in this study.

\section*{Funding}
\begin{itemize}
    \item Federal Institute of Education, Science and Technology of Pará - IFPA;
    
    \item Coordination for the Improvement of Higher Education Personnel - Brazil (CAPES) [Financing Code 001];
    
    \item Vale Institute of Technology - ITV: [grant number R100603.DTL.07] and  [grant number R100603.DTL.08];
    
\end{itemize}

\section*{Conflicts of interest/Competing interests}

It is declared that there are no conflicts of interest between the authors and their institutions belonging to any part of this research.

\section*{Ethics approval}

There is no need for the approval of a research ethics council, because the personal identity of the consulted experts is not presented, because they work only with the quantitative aspects of their answers and because the data consulted are in the public domain obtained through request to the responsible institution.

\section*{Consent to participate}

All individuals involved (researchers and volunteers) in this research declare that they are aware of its nature and seriousness, together with its importance to society.

\section*{Consent for publication}
All individuals and institutions involved in the research in question are in agreement with the publication of this article in the journal.

\section*{Availability of data and material}\label{lab_availability}

\begin{itemize}
    \item[A] - All data used in this study are available at: 

    \begin{itemize}
    \item[A.1] - \url{https://github.com/josesousaribeiro/Pred2Town-and-XAI};
    \item[A.2] - \url{https://github.com/josesousaribeiro/Pred2Town}.
    \end{itemize}

    \item[B] - The form applied to public security specialists can be accessed at:
    \begin{itemize}
    \item[B.1] - \url{https://sites.google.com/view/survey-peridico-pred2town/in%C3%ADcio}
    \end{itemize}
\end{itemize}

\section*{Code availability} \label{lab_code} 

\begin{itemize}
\item[A] - All code used in this study are available at: 
\begin{itemize}
    \item[A.1] - \url{https://github.com/josesousaribeiro/Pred2Town-and-XAI/blob/main/XAI_Pred2Town.ipynb};
    \item[A.2] - \url{https://github.com/josesousaribeiro/Pred2Town/blob/master/Notebook_Pred2Town_XAI_v1.1.ipynb};
    \item[A.3] - \url{https://github.com/josesousaribeiro/ConeXi} \label{lab_conexi}
\end{itemize}
\end{itemize}

\section*{Authors' contributions}

All authors contributed to the conception and design of the study in general. Material preparation, data collection and analysis were carried out by José Ribeiro. The revisions to the writing and design of results visualizations were made by Níkolas Carneiro. The writing revisions and statistical analyzes were done by Lucas Cardoso. All activities were carried out under the supervision and planning of Ronnie Alves.

%
%
%
\bibliographystyle{splncs04}
\bibliography{main}

\begin{thebibliography}{10}
\providecommand{\url}[1]{\texttt{#1}}
\providecommand{\urlprefix}{URL }
\providecommand{\doi}[1]{https://doi.org/#1}

\bibitem{eli5_git}
Eli5 {Git}. {https://github.com/TeamHG-Memex/eli5}

\bibitem{ethical_ml_git}
{Ethical ML}.
  https://github.com/EthicalML/awesome-production-machine-learning\#explaining-black-box-models-and-datasets,
  \hfill Last \hfill accessed \hfill 20 \hfill Apr \hfill 2021.

\bibitem{scikit-learn}
Scikit-learn. {https://scikit-learn.org/0.22/}, \hfill Last \hfill accessed
  \hfill 21 \hfill Nov \hfill 2020.

\bibitem{skater_ref}
Skater. {https://www.oreilly.com/content/interpreting-predictive-models-with
  -skater-unboxing-model-opacity/}, \hfill Last \hfill accessed \hfill 21
  \hfill Jan \hfill 2021.

\bibitem{skater_git}
Skater {Git}. https://github.com/oracle/Skater, \hfill Last \hfill accessed
  \hfill 21 \hfill Jan \hfill 2021.

\bibitem{al2019_crime_1}
AL~Mansour, H., Lundy, M.: Crime types prediction. In: Advances in Data
  Science, Cyber Security and IT Applications: First International Conference
  on Computing, ICC 2019, Riyadh, Saudi Arabia, December 10--12, 2019,
  Proceedings, Part I 1. pp. 260--274. Springer (2019)

\bibitem{alderden2007_crime_8}
Alderden, M.A., Lavery, T.A.: Predicting homicide clearances in chicago:
  Investigating disparities in predictors across different types of homicide.
  Homicide Studies  \textbf{11}(2),  115--132 (2007)

\bibitem{ang2015san_crime_2}
Ang, S.T., Wang, W., Chyou, S.: San francisco crime classification. University
  of California San Diego  (2015)

\bibitem{angerschmid2022fairness}
Angerschmid, A., Zhou, J., Theuermann, K., Chen, F., Holzinger, A.: Fairness
  and explanation in ai-informed decision making. Machine Learning and
  Knowledge Extraction  \textbf{4}(2),  556--579 (2022)

\bibitem{alibi_ale_ref}
Apley, D.W., Zhu, J.: Visualizing the effects of predictor variables in black
  box supervised learning models. Journal of the Royal Statistical Society:
  Series B (Statistical Methodology)  \textbf{82}(4),  1059--1086 (2020)

\bibitem{spearman_ref}
Artusi, R., Verderio, P., Marubini, E.: Bravais-{Pearson} and {Spearman}
  {Correlation} {Coefficients}: {Meaning}, {Test} of {Hypothesis} and
  {Confidence} {Interval}. The International Journal of Biological Markers
  \textbf{17}(2),  148--151 (Apr 2002),
  \url{https://doi.org/10.1177/172460080201700213}, publisher: SAGE
  Publications Ltd STM

\bibitem{artusi2002bravais_confidence}
Artusi, R., Verderio, P., Marubini, E.: Bravais-pearson and spearman
  correlation coefficients: meaning, test of hypothesis and confidence
  interval. The International journal of biological markers  \textbf{17}(2),
  148--151 (2002)

\bibitem{ibm_xai360}
Arya, V., Bellamy, R.K., Chen, P.Y., Dhurandhar, A., Hind, M., Hoffman, S.C.,
  Houde, S., Liao, Q.V., Luss, R., Mojsilovic, A., et~al.: Ai explainability
  360: An extensible toolkit for understanding data and machine learning
  models. J. Mach. Learn. Res.  \textbf{21}(130), ~1--6 (2020)

\bibitem{dalex_python_ref}
Baniecki, H., Kretowicz, W., Piatyszek, P., Wisniewski, J., Biecek, P.: dalex:
  Responsible machine learning with interactive explainability and fairness in
  python. Journal of Machine Learning Research  \textbf{22}(214), ~1--7 (2021),
  \url{http://jmlr.org/papers/v22/20-1473.html}

\bibitem{arrieta_explainable_2019_20}
Barredo~Arrieta, A., Díaz-Rodríguez, N., Del~Ser, J., Bennetot, A., Tabik,
  S., Barbado, A., Garcia, S., Gil-Lopez, S., Molina, D., Benjamins, R.,
  Chatila, R., Herrera, F.: Explainable {Artificial} {Intelligence} ({XAI}):
  {Concepts}, taxonomies, opportunities and challenges toward responsible {AI}.
  Information Fusion  \textbf{58},  82--115 (Jun 2020).
  \doi{10.1016/j.inffus.2019.12.012}

\bibitem{dalex_r_ref}
Biecek, P.: Dalex: Explainers for complex predictive models in r. Journal of
  Machine Learning Research  \textbf{19}(84), ~1--5 (2018),
  \url{https://jmlr.org/papers/v19/18-416.html}

\bibitem{bishara2017confidence_conf_notnormal}
Bishara, A.J., Hittner, J.B.: Confidence intervals for correlations when data
  are not normal. Behavior research methods  \textbf{49},  294--309 (2017)

\bibitem{breiman2001random_ML_randomforest}
Breiman, L.: Random forests. Machine learning  \textbf{45}(1),  5--32 (2001)

\bibitem{dalex_book}
Burzykowski, P.B.a.T.: Explanatory {Model} {Analysis}.
  \url{https://ema.drwhy.ai/}

\bibitem{teoria_crime_cano_as_2002}
Cano, I., Soares, G.D.: As teorias sobre as causas da criminalidade  (2002)

\bibitem{lofo_ref}
{\c{C}}ay{\i}r, U., Yenido{\u{g}}an, I., Da{\u{g}}, H.: Use case study: Data
  science application for microsoft malware prediction competition on kaggle.
  Proceedings Book p.~98 (2019)

\bibitem{cerqueira_crimes_scielo}
Cerqueira, D., Lob{\~a}o, W.: Determinantes da criminalidade: arcabou{\c{c}}os
  te{\'o}ricos e resultados emp{\'\i}ricos. Dados  \textbf{47},  233--269
  (2004)

\bibitem{chadaga2023artificial_compare}
Chadaga, K., Prabhu, S., Bhat, V., Sampathila, N., Umakanth, S., Chadaga, R.:
  Artificial intelligence for diagnosis of mild--moderate covid-19 using
  haematological markers. Annals of Medicine  \textbf{55}(1),  2233541 (2023)

\bibitem{xgb}
Chen, T., He, T., Benesty, M., Khotilovich, V., Tang, Y., Cho, H., Chen, K.,
  Mitchell, R., Cano, I., Zhou, T., et~al.: Xgboost: extreme gradient boosting.
  R package version 0.4-2  \textbf{1}(4), ~1--4 (2015)

\bibitem{colasanti2023homicide_crime_9}
Colasanti, M., Ricci, E., Cardinale, A., Amati, F., Mazza, C., Biondi, S.,
  Ferracuti, S., Roma, P.: Homicide-suicide in italy between 2009-2018: An
  epidemiological update and time series analysis. European Journal on Criminal
  Policy and Research pp. 1--16 (2023)

\bibitem{cortez2011opening_blckbox_opening}
Cortez, P., Embrechts, M.J.: Opening black box data mining models using
  sensitivity analysis. In: 2011 IEEE Symposium on Computational Intelligence
  and Data Mining (CIDM). pp. 341--348. IEEE (2011)

\bibitem{teoria_crime_1968}
Cressey, D.P.: Crime: {Causes} of {Crime} in {International} {Encyclopedia} of
  the {Social} {Sciences}  (1968)

\bibitem{cb}
Dorogush, A.V., Ershov, V., Gulin, A.: Catboost: gradient boosting with
  categorical features support. arXiv preprint arXiv:1810.11363  (2018)

\bibitem{skater_citation}
Dwivedi, R., Dave, D., Naik, H., Singhal, S., Omer, R., Patel, P., Qian, B.,
  Wen, Z., Shah, T., Morgan, G., et~al.: Explainable ai (xai): Core ideas,
  techniques, and solutions. ACM Computing Surveys  \textbf{55}(9),  1--33
  (2023)

\bibitem{ethical_xai_site}
for Ethical~AI, T.I., Learning, M.: Ethical {XAI} (2021),
  \url{https://ethical.institute/xai.html}, \hfill Last \hfill accessed \hfill
  22 \hfill Jul \hfill 2021.

\bibitem{lgb}
Fan, J., Ma, X., Wu, L., Zhang, F., Yu, X., Zeng, W.: Light gradient boosting
  machine: An efficient soft computing model for estimating daily reference
  evapotranspiration with local and external meteorological data. Agricultural
  water management  \textbf{225},  105758 (2019)

\bibitem{ferreira2015does_pvalue}
Ferreira, J.C., Patino, C.M.: What does the p value really mean? Jornal
  Brasileiro de Pneumologia  \textbf{41}(5), ~485 (2015)

\bibitem{ciu_ref}
Fr{\"a}mling, K.: Decision theory meets explainable ai. In: International
  Workshop on Explainable, Transparent Autonomous Agents and Multi-Agent
  Systems. pp. 57--74. Springer (2020)

\bibitem{ghahramani2015probabilistic}
Ghahramani, Z.: Probabilistic machine learning and artificial intelligence.
  Nature  \textbf{521}(7553),  452--459 (2015)

\bibitem{teoria_contra_pessoas_gilaberte2013crimes}
GILABERTE, B.: Crimes contra a pessoa. Rio de Janeiro: Freitas Bastos  (2013)

\bibitem{interpretabilidade_review2018}
Gilpin, L.H., Bau, D., Yuan, B.Z., Bajwa, A., Specter, M., Kagal, L.:
  Explaining explanations: An overview of interpretability of machine learning.
  In: 2018 IEEE 5th International Conference on Data Science and Advanced
  Analytics (DSAA). pp. 80--89 (2018). \doi{10.1109/DSAA.2018.00018}

\bibitem{expectation_ronny}
Goldstein, A., Kapelner, A., Bleich, J., Pitkin, E.: Peeking inside the black
  box: Visualizing statistical learning with plots of individual conditional
  expectation. Journal of Computational and Graphical Statistics
  \textbf{24}(1),  44--65 (2015). \doi{10.1080/10618600.2014.907095},
  \url{https://doi.org/10.1080/10618600.2014.907095}

\bibitem{guidotti2018survey}
Guidotti, R., Monreale, A., Ruggieri, S., Turini, F., Giannotti, F., Pedreschi,
  D.: A survey of methods for explaining black box models. ACM computing
  surveys (CSUR)  \textbf{51}(5),  1--42 (2018)

\bibitem{darpa_2019}
Gunning, D., Aha, D.: {DARPA}’s {Explainable} {Artificial} {Intelligence}
  ({XAI}) {Program}. AI Magazine  \textbf{40}(2),  44--58 (Jun 2019).
  \doi{10.1609/aimag.v40i2.2850},
  \url{https://ojs.aaai.org/index.php/aimagazine/article/view/2850}, number: 2

\bibitem{gunning2019xai}
Gunning, D., Stefik, M., Choi, J., Miller, T., Stumpf, S., Yang, G.Z.:
  Xai—explainable artificial intelligence. Science Robotics  \textbf{4}(37),
  eaay7120 (2019)

\bibitem{hall2017machine}
Hall, P., Gill, N., Kurka, M., Phan, W.: Machine learning interpretability with
  h2o driverless ai. H2O. ai  (2017)

\bibitem{hariharan2023xai_compare}
Hariharan, S., Rejimol~Robinson, R., Prasad, R.R., Thomas, C., Balakrishnan,
  N.: Xai for intrusion detection system: comparing explanations based on
  global and local scope. Journal of Computer Virology and Hacking Techniques
  \textbf{19}(2),  217--239 (2023)

\bibitem{holzinger2020explainable}
Holzinger, A., Saranti, A., Molnar, C., Biecek, P., Samek, W.: Explainable ai
  methods-a brief overview. In: International Workshop on Extending Explainable
  AI Beyond Deep Models and Classifiers. pp. 13--38. Springer (2020)

\bibitem{hong2020human}
Hong, S.R., Hullman, J., Bertini, E.: Human factors in model interpretability:
  Industry practices, challenges, and needs. Proceedings of the ACM on
  Human-Computer Interaction  \textbf{4}(CSCW1),  1--26 (2020)

\bibitem{ibge_belem}
{IBGE}: {https://cidades.ibge.gov.br/brasil/pa/belem/panorama}, \hfill Last
  \hfill accessed \hfill 04 \hfill Mar \hfill 2022.

\bibitem{ipea_mapa_violencia}
{IPEA}:
  {www.ipea.gov.br/atlasviolencia/arquivos/downloads/7047-190802atlasdaviolencia2019municipios.pdf},
  \hfill Last \hfill accessed \hfill 04 \hfill Mar \hfill 2021.

\bibitem{jin2020addressing_crime_4}
Jin, G., Wang, Q., Zhu, C., Feng, Y., Huang, J., Zhou, J.: Addressing crime
  situation forecasting task with temporal graph convolutional neural network
  approach. In: 2020 12th International Conference on Measuring Technology and
  Mechatronics Automation (ICMTMA). pp. 474--478. IEEE (2020)

\bibitem{jouis2021_anchors}
Jouis, G., Mouch{\`e}re, H., Picarougne, F., Hardouin, A.: Anchors vs
  attention: Comparing xai on a real-life use case. In: Del~Bimbo, A.,
  Cucchiara, R., Sclaroff, S., Farinella, G.M., Mei, T., Bertini, M.,
  Escalante, H.J., Vezzani, R. (eds.) Pattern Recognition. ICPR International
  Workshops and Challenges. pp. 219--227. Springer International Publishing,
  Cham (2021)

\bibitem{ciu_git}
Kampik, T.: Ciu {Git} (Mar 2021), \url{https://github.com/TimKam/py-ciu},
  original-date: 2020-01-15T08:52:28Z

\bibitem{kaur2020interpreting}
Kaur, H., Nori, H., Jenkins, S., Caruana, R., Wallach, H., Wortman~Vaughan, J.:
  Interpreting interpretability: Understanding data scientists' use of
  interpretability tools for machine learning. In: Proceedings of the 2020 CHI
  Conference on Human Factors in Computing Systems. pp. 1--14 (2020)

\bibitem{decisions_1993}
Keeney, R.L., L, K.R., Howard, R.: Decisions with {Multiple} {Objectives}:
  {Preferences} and {Value} {Trade}-{Offs}. Cambridge University Press,
  Cambridge England ; New York, NY, USA, revised ed. edição edn. (Aug 1993)

\bibitem{khan2022model_specific}
Khan, A.: Model-specific explainable artificial intelligence techniques:
  State-of-the-art, advantages and limitations  (2022)

\bibitem{kindermans_learning_2017_2018}
Kindermans, P.J., Schütt, K.T., Alber, M., Müller, K.R., Erhan, D., Kim, B.,
  Dähne, S.: Learning how to explain neural networks: Patternnet and
  patternattribution. In: International Conference on Learning Representations
  (2018), \url{https://openreview.net/forum?id=Hkn7CBaTW}

\bibitem{dt}
Kingsford, C., Salzberg, S.L.: What are decision trees? Nature biotechnology
  \textbf{26}(9),  1011--1013 (2008)

\bibitem{machinelearning_ML_safety}
Kiran, M., Khan, S.: Machine learning for safety--critical applications in
  engineering. Machine Learning  \textbf{109}(5),  1101--1102 (2020)

\bibitem{eli5_ref}
Korobov, M., Lopuhin, K.: Eli5.
  {https://eli5.readthedocs.io/en/latest/index.html}, \hfill Last \hfill
  accessed \hfill 21 \hfill Jan \hfill 2021.

\bibitem{lr}
LaValley, M.P.: Logistic regression. Circulation  \textbf{117}(18),  2395--2399
  (2008)

\bibitem{review_xai_2021}
Linardatos, P., Papastefanopoulos, V., Kotsiantis, S.: Explainable ai: A review
  of machine learning interpretability methods. Entropy  \textbf{23}(1) (2021).
  \doi{10.3390/e23010018}, \url{https://www.mdpi.com/1099-4300/23/1/18}

\bibitem{shap_ref}
Lipovetsky, S., Conklin, M.: Analysis of regression in game theory approach.
  Applied Stochastic Models in Business and Industry  \textbf{17}(4),  319--330
  (2001)

\bibitem{svm}
Lorena, A.C., De~Carvalho, A.C.: Uma introdu{\c{c}}{\~a}o {\`a}s support vector
  machines. Revista de Inform{\'a}tica Te{\'o}rica e Aplicada  \textbf{14}(2),
  43--67 (2007)

\bibitem{shap_doc}
Lundberg, S.: Shap {Documentation} (2023),
  \url{https://shap.readthedocs.io/en/latest/}, original-date:
  2023-01-15T08:52:28Z

\bibitem{xai_local_global_2020}
Lundberg, S.M., Erion, G., Chen, H., DeGrave, A., Prutkin, J.M., Nair, B.,
  Katz, R., Himmelfarb, J., Bansal, N., Lee, S.I.: From local explanations to
  global understanding with explainable {AI} for trees. Nature Machine
  Intelligence  \textbf{2}(1),  56--67 (Jan 2020).
  \doi{10.1038/s42256-019-0138-9},
  \url{https://www.nature.com/articles/s42256-019-0138-9}, number: 1 Publisher:
  Nature Publishing Group

\bibitem{tree_shap_ref}
Lundberg, S.M., Erion, G., Chen, H., DeGrave, A., Prutkin, J.M., Nair, B.,
  Katz, R., Himmelfarb, J., Bansal, N., Lee, S.I.: From local explanations to
  global understanding with explainable ai for trees. Nature Machine
  Intelligence  \textbf{2}(1),  2522--5839 (2020)

\bibitem{damasceno2011prediction_crime_3}
Damasceno~de Melo, M., Teixeira, J., Campos, G.: A prediction model for
  criminal levels specialized in brazilian cities. In: International Conference
  on e-Democracy. pp. 131--138. Springer (2011)

\bibitem{interpretabilidade_messalas2019}
Messalas, A., Kanellopoulos, Y., Makris, C.: Model-agnostic interpretability
  with shapley values. In: 2019 10th International Conference on Information,
  Intelligence, Systems and Applications (IISA). pp.~1--7 (2019).
  \doi{10.1109/IISA.2019.8900669}

\bibitem{molnar2020interpretable}
Molnar, C.: Interpretable machine learning. Lulu. com (2020)

\bibitem{muller2021ten}
Muller, H., Mayrhofer, M.T., Van~Veen, E.B., Holzinger, A.: The ten
  commandments of ethical medical ai. Computer  \textbf{54}(07),  119--123
  (2021)

\bibitem{gb}
Natekin, A., Knoll, A.: Gradient boosting machines, a tutorial. Frontiers in
  neurorobotics  \textbf{7}, ~21 (2013)

\bibitem{human_in_the_loop_zanzoto}
Nguyen, T.N., Choo, R.: Human-in-the-loop xai-enabled vulnerability detection,
  investigation, and mitigation. In: 2021 36th IEEE/ACM International
  Conference on Automated Software Engineering (ASE). pp. 1210--1212. IEEE
  (2021)

\bibitem{interpretML_arxiv}
Nori, H., Jenkins, S., Koch, P., Caruana, R.: Interpretml: A unified framework
  for machine learning interpretability. arXiv preprint arXiv:1909.09223
  (2019)

\bibitem{sou_da_paz}
da~Paz, I.S.: Onde mora a impunidade?

\bibitem{knn}
Peterson, L.E.: K-nearest neighbor. Scholarpedia  \textbf{4}(2), ~1883 (2009)

\bibitem{explicabilidade_rede_neural}
Qi, Z., Khorram, S., Li, F.: Visualizing deep networks by optimizing with
  integrated gradients. In: CVPR Workshops. vol.~2 (2019)

\bibitem{info_theory_1994}
Reza, F.M.: An {Introduction} to {Information} {Theory}. Courier Corporation
  (Jan 1994), google-Books-ID: RtzpRAiX6OgC

\bibitem{ribeiro_pred2town_et_al_2021}
Ribeiro, J., Meneses, L., Costa, D., Miranda, W., Alves, R.: Prediction of
  homicides in urban centers: A machine learning approach. In: Arai, K. (ed.)
  Intelligent Systems and Applications. pp. 344--361. Springer International
  Publishing, Cham (2022)

\bibitem{ribeiro_complexity_et_al_2021}
Ribeiro, J., Silva, R., Cardoso, L., Alves, R.: Does dataset complexity matters
  for model explainers? In: 2021 IEEE International Conference on Big Data (Big
  Data). pp. 5257--5265 (2021). \doi{10.1109/BigData52589.2021.9671630}

\bibitem{lime_ref}
Ribeiro, M.T., Singh, S., Guestrin, C.: "why should {I} trust you?": Explaining
  the predictions of any classifier. In: Proceedings of the 22nd {ACM} {SIGKDD}
  International Conference on Knowledge Discovery and Data Mining, San
  Francisco, CA, USA, August 13-17, 2016. pp. 1135--1144 (2016)

\bibitem{ribeiro2018anchors}
Ribeiro, M.T., Singh, S., Guestrin, C.: Anchors: High-precision model-agnostic
  explanations. In: Proceedings of the AAAI conference on artificial
  intelligence. vol.~32 (2018)

\bibitem{roth1988shapley}
Roth, A.E.: The Shapley value: essays in honor of Lloyd S. Shapley. Cambridge
  University Press (1988)

\bibitem{sahatova2022overview_comparacao_shap_lime}
Sahatova, K., Balabaeva, K.: An overview and comparison of xai methods for
  object detection in computer tomography. Procedia Computer Science
  \textbf{212},  209--219 (2022)

\bibitem{shalev2014understanding}
Shalev-Shwartz, S., Ben-David, S.: Understanding machine learning: From theory
  to algorithms. Cambridge university press (2014)

\bibitem{shermila2018_crime_5}
Shermila, A.M., Bellarmine, A.B., Santiago, N.: Crime data analysis and
  prediction of perpetrator identity using machine learning approach. In: 2018
  2nd international conference on trends in electronics and informatics
  (ICOEI). pp. 107--114. IEEE (2018)

\bibitem{interpretabilidade_singh2020model}
Singh, J., Anand, A.: Model agnostic interpretability of rankers via intent
  modelling. In: Proceedings of the 2020 Conference on Fairness,
  Accountability, and Transparency. pp. 618--628 (2020)

\bibitem{sokol2020explainability_peterflash_humanintheloop}
Sokol, K., Flach, P.: Explainability fact sheets: a framework for systematic
  assessment of explainable approaches. In: Proceedings of the 2020 Conference
  on Fairness, Accountability, and Transparency. pp. 56--67 (2020)

\bibitem{peter2020explainability}
Sokol, K., Flach, P.: Explainability fact sheets: a framework for systematic
  assessment of explainable approaches. In: Proceedings of the 2020 Conference
  on Fairness, Accountability, and Transparency. pp. 56--67 (2020)

\bibitem{peterflach_humanintheloop}
Sokol, K., Flach, P.: One explanation does not fit all. KI-K{\"u}nstliche
  Intelligenz  \textbf{34}(2),  235--250 (2020)

\bibitem{escala_ordinal_estatistica}
Stevens, S.S., et~al.: On the theory of scales of measurement  (1946)

\bibitem{nn}
Wang, S.C., Wang, S.C.: Artificial neural network. Interdisciplinary computing
  in java programming pp. 81--100 (2003)

\bibitem{nb}
Webb, G.I., Keogh, E., Miikkulainen, R.: Na{\"\i}ve bayes. Encyclopedia of
  machine learning  \textbf{15}(1),  713--714 (2010)

\bibitem{wenninger2022explainable_qlattice}
Wenninger, S., Kaymakci, C., Wiethe, C.: Explainable long-term building energy
  consumption prediction using qlattice. Applied Energy  \textbf{308},  118300
  (2022)

\bibitem{machine_learning_ML_health}
Wiens, J., Wallace, B.C.: special issue on machine learning for health and
  medicine. Machine Learning  \textbf{102}(3),  305--307 (2016)

\bibitem{wilming2022scrutinizing_ML_human_in_the_loop}
Wilming, R., Budding, C., M{\"u}ller, K.R., Haufe, S.: Scrutinizing xai using
  linear ground-truth data with suppressor variables. Machine learning pp.
  1--21 (2022)

\bibitem{yadav2017_crime_7}
Yadav, S., Timbadia, M., Yadav, A., Vishwakarma, R., Yadav, N.: Crime pattern
  detection, analysis \& prediction. In: 2017 International conference of
  Electronics, Communication and Aerospace Technology (ICECA). vol.~1, pp.
  225--230. IEEE (2017)

\end{thebibliography}

\end{document}